%% file: main.tex
\PassOptionsToPackage{shortlabels}{enumitem}
\documentclass[11pt, a4paper]{include/gdm_format}
\usepackage[authoryear, sort&compress, round]{natbib}
\usepackage{xcolor}
\usepackage{subcaption}
\usepackage{mwe}
\usepackage{graphicx}
\usepackage{xr}
\usepackage{marvosym}
\usepackage{makecell}
\usepackage{graphicx}

\usepackage{fvextra}
\usepackage{tabularray}
\UseTblrLibrary{booktabs}
\definecolor{fugublue}{HTML}{EAF2FB}
\usepackage{graphicx}
\newcommand{\best}[1]{\scalebox{1.05}{\textbf{#1}}}   
\newcommand{\second}[1]{\underline{#1}} 

\definecolor{bgcolor}{rgb}{0.95,0.95,0.95}

\definecolor{lblA}{RGB}{33,113,181}
\definecolor{lblB}{RGB}{230,85,13}

\usepackage{tablefootnote}
\usepackage{pdfpages}
\usepackage{etoc}

\usepackage{hyperref}
\usepackage{url}
\usepackage{xurl}
\usepackage[parfill]{parskip}

\usepackage{amsmath}
\usepackage{xspace}
\usepackage{multirow}
\usepackage{paracol}
\usepackage{booktabs}
\usepackage{color}
\usepackage{colortbl}
\usepackage{cleveref}
\usepackage{graphicx}
\usepackage{algorithm}
\usepackage{algorithmicx}
\usepackage{algpseudocode}
\usepackage{subcaption}
\usepackage{wrapfig}
\usepackage{sidecap}
\usepackage{soul}
\sidecaptionvpos{figure}{t}
\usepackage{multicol}
\usepackage{pifont}
\usepackage[normalem]{ulem}
\usepackage{listings}
\usepackage{changes}

\title{Sakana Fugu Technical Report}

\author{Fugu Team, Sakana AI \footnote{Please cite this work as "Sakana AI (2026)". Full authorship and contributions appear at the end of the document.}}

\usepackage{tcolorbox}
\tcbuselibrary{breakable}
\tcbuselibrary{listings}  
\definecolor{lightgreen}{RGB}{235, 255, 235}  

\lstdefinelanguage{CUDA}{
    morekeywords={
        __global__, __device__, __host__, __shared__, __constant__,
        dim3, gridDim, blockDim, blockIdx, threadIdx, syncthreads
    },
    sensitive=true,
    morecomment=[l]{//},
    morecomment=[s]{/*}{*/},
    morestring=[b]"
}

\definecolor{codegreen}{rgb}{0,0.6,0}
\definecolor{codegray}{rgb}{0.5,0.5,0.5}
\definecolor{codepurple}{rgb}{0.58,0,0.82}
\definecolor{backcolour}{RGB}{245,248,250}
\definecolor{emph}{RGB}{166,88,53}
\definecolor{nightblue}{RGB}{9,49,105}
\definecolor{keywords}{RGB}{207,33,46}
\definecolor{lightpurple}{RGB}{130,81,223}

\lstdefinestyle{mystyle}{
    backgroundcolor=\color{backcolour},   
    commentstyle=\color{codegreen},
    keywordstyle=\color{keywords},
    stringstyle=\color{nightblue},
    basicstyle=\ttfamily\scriptsize,
    breakatwhitespace=false,         
    breaklines=true,                 
    captionpos=b,                    
    keepspaces=true,                 
    showspaces=false,                
    showstringspaces=false,
    showtabs=false,                  
    tabsize=2,
    frame=shadowbox,
    emph={AutoTokenizer,AutoModelForSequenceClassification,Explainer},
    emphstyle={\color{emph}},
    emph={[2]from_pretrained,compute_table},
    emphstyle={[2]\color{lightpurple}},
}

\usepackage{CJKutf8}
\begin{document}

\begin{abstract}
\input{sections/00_abstract.tex}
\end{abstract}

\maketitle
\setcounter{tocdepth}{2}
\etocdepthtag.toc{mtchapter}
\etocsettagdepth{mtchapter}{subsection}
\etocsettagdepth{mtappendix}{none}

\section{Introduction}

\input{sections/01_introduction.tex}

\section{Related Work}

\input{sections/02_background.tex}

\section{Sakana Fugu}
\label{sec:sakana_fugu}
\input{sections/03_sakana_fugu}
\input{sections/03_fugumini.tex}
\input{sections/03_fuguultra.tex}


\section{Capabilities}
\label{sec:results}

\input{sections/04_results.tex}


\section{Conclusions}
\input{sections/06_conclusion.tex}

\section*{Acknowledgments}
We thank all Sakana AI members who adopted Fugu models in their daily workflows and helped shape the system through continuous feedback.
We are also deeply grateful to our external early users, whose detailed bug reports, thoughtful suggestions, and demanding real-world use cases were invaluable in improving Sakana Fugu.

\section*{Authors List and Technical Contributions}
\label{sec:author_contribution_list}

Team members are listed alphabetically by their first name.
As a small team, contributions naturally spanned many areas, and most members helped well beyond any single role.
We list only the main contributions here.

\noindent\textbf{Team \& Project Lead:}
Yujin Tang

\noindent\textbf{Core Contributors (model):}
Edoardo Cetin, Jinglue Xu, Qi Sun, Stefan Nielsen, Vincent Richard

\noindent\textbf{Core Contributors (infrastructure):}
Haruto Goda, Iaroslav Tymchenko, Nhan Nguyen

\noindent\textbf{Contributors:}
Hyunin Lee, Mari Ashiga, Shashank Kotyan, So Kuroki, Tarin Clanuwat

\newpage
\bibliography{references}
\bibliographystyle{plainnat}

\clearpage
\appendix

\etocdepthtag.toc{mtappendix}
\etocsettagdepth{mtchapter}{none}
\etocsettagdepth{mtappendix}{subsection}
\renewcommand{\contentsname}{Appendix}

\newpage
\input{sections/appendix.tex}

\end{document}

%% file: sections/00_abstract.tex
The capabilities of frontier Large Language Models (LLMs) continue to advance, with different providers increasingly specializing in distinct domains.
This raises a natural next objective: how to combine the individual specializations of various LLMs into a collectively intelligent system.
To this end, we report the development of Sakana Fugu, a family of orchestrator models that harness and amplify the capabilities of an LLM agent team.
Fugu models are themselves language models trained to understand user queries and dynamically devise agentic scaffolds to solve them.
Through these adaptive scaffolds, Fugu accesses performance beyond any individual LLM agent, achieving state-of-the-art results compared to other publicly accessible models across a range of challenging tasks, including SWE-Bench Pro, Terminal Bench, LiveCodeBench, GPQA-Diamond, Humanity's Last Exam, and CharXiv Reasoning. 
We release two models: \textsc{Fugu}, which balances performance with latency for everyday use, and \textsc{Fugu-Ultra}, which prioritizes answer quality on the hardest problems.
We describe our training paradigm, which encompasses large-scale fine-tuning, evolutionary algorithms, and reinforcement learning approaches, along with the infrastructure and core design principles that turn these methods into a production system.
We hope this report encourages further research into multi-agent systems and dynamic, query-adaptive agentic scaffolds as a path toward the next frontier of AI capabilities, accessed through collective intelligence.
\url{https://sakana.ai/fugu}

%% file: sections/01_introduction.tex
\begin{figure}[!ht]
    \centering
    \includegraphics[width=1.0\textwidth]{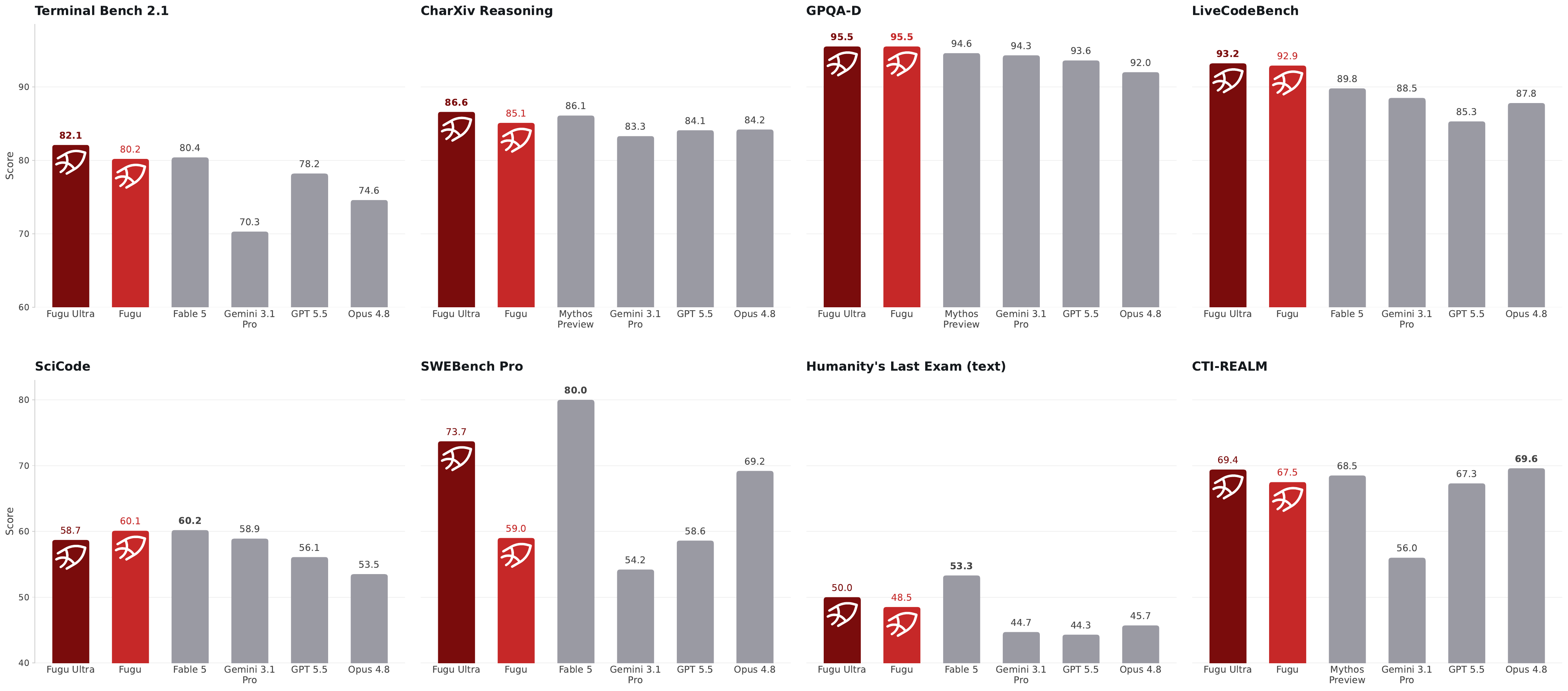}
    \caption{Performance comparison of \textsc{Fugu}, \textsc{Fugu-Ultra}, and baseline frontier models across a suite of coding, reasoning, scientific, and agentic benchmarks. All scores other than Fugu's are reported by the model providers. For Fable 5 and Mythos Preview, we report the max of the two if both scores are available on the same benchmark. Neither of them is in Fugu's agent pool as they are not publicly accessible. }
    \label{fig:cover}
\end{figure}

Frontier Large Language Models (LLMs) now approach or match expert-level performance across a wide range of domains \citep{luong2026imo, anthropic2026chemist, openai2026unitdistance, patwardhan2025gdpval, nori2025sequential}.
For example, Gemini-3-Deep-Think recently achieved gold-medal performance at the International Mathematical Olympiad \citep{luong2026imo}, GPT-5.5 disproved an 80-year-old Erd\H{o}s conjecture in combinatorial geometry \citep{openai2026unitdistance}, and Claude-Mythos uncovered numerous zero-day vulnerabilities in OpenBSD and FreeBSD \citep{carlini2026mythos}.
Yet as frontier capabilities advance, we also observe increasingly differentiated model specializations at multiple levels of granularity.
At the domain level, past GPT-series models often displayed state-of-the-art performance on mathematical reasoning tasks \citep{openai2026unitdistance, glazer2024frontiermath}, while Opus-series models have specialized in software engineering and cybersecurity \citep{carlini2026mythos, jimenez2024swebench}.
Within a single domain, the differences can be even more fine-grained: in competitive coding, Gemini-3.1-Pro can be particularly effective at directly implementing known algorithms, while GPT-series models can often excel at planning and combining multiple algorithmic ideas to solve the hardest problems \citep{zheng2026livecodebench}.
These observations suggest that the next frontier may not be achieved by any single model alone, but by systems that can identify, combine, and amplify the complementary strengths of many models.
A system that can engineer software like the strongest coding model, reason mathematically like the strongest theorem-proving model, and dynamically decide when each specialization is needed would provide a natural path toward expanding frontier performance through collective intelligence.

At the same time, recent improvements in LLM performance have been driven not only by advances in the underlying models but also by agentic scaffolds that treat the LLM as one component within a larger system \citep{hu2025automated}.
Such scaffolds augment autoregressive generation with structured prompting \citep{wei2022chain, yao2022react, madaan2023self}, external tool use and function calling \citep{qu2025tool,schick2024toolformer}, environment feedback, and memory management \citep{zhang2025survey}.
This is especially visible in software engineering, where Claude Code and other modern agentic scaffolds can shape the model's interaction trajectory, supply continuous environment feedback, and turn raw model capability into an iterative agentic workflow \citep{anthropic_claude_code}.
These developments suggest that capability should be understood not only as a property of the model, but also as a property of the scaffold through which the model acts.

The growing range of language models with diverse capabilities and the impact of domain-specific agentic scaffolds motivate the development of systems that can orchestrate collective intelligence: dynamically choosing which models to involve, how they should communicate, use tools and interact with the environment, and when their outputs should be synthesized. We view this kind of model orchestration as a new complementary scaling axis beyond ever larger and expensive language models. If successful, this direction could make frontier-level capability more efficient, more modular, and more broadly accessible.

In this technical report, we detail the development of Sakana Fugu, our system harnessing the benefits of collective intelligence to realize this vision.
Sakana Fugu is a family of language models trained to adaptively and dynamically orchestrate a team of more powerful frontier agent workers.
In our initial release, we make two variants available to the public, targeting different operating points.
\textsc{\textbf{Fugu}} is optimized for speed, selecting a single worker per input so that its latency is comparable to a direct call to a frontier model, while still routing each query to the most capable agent for that input.
Despite this efficiency, \textsc{Fugu} matches and in many cases surpasses best-in-class performance across tasks.
\textsc{\textbf{Fugu-Ultra}} is optimized for performance, composing workflows of multiple agents per input.
It therefore trades additional latency for higher quality and is intended for the most complex tasks that benefit from combining multiple specializations.
Figure~\ref{fig:cover} compares Sakana Fugu with frontier models on popular benchmarks.
Our Fugu models surpass publicly accessible frontier models and are shoulder-to-shoulder with Fable 5 and Mythos Preview in various rigorous engineering, scientific, and reasoning benchmarks while delivering frontier capability without the risk of export controls.
See section~\ref{sec:results} for more results.

Beyond raw performance, we believe learned orchestration carries broader implications.
If capability can be amplified by composing existing models rather than only by training larger ones, then progress in AI need not depend solely on access to the largest training runs. 
Orchestration offers a more modular and accessible path, in which improvements arise from how models are selected, coordinated, and combined, and in which newly released models can be incorporated into the worker pool as they appear.
This composability also gives users meaningful control over the system.
For example, agent pools can be configured to favor particular providers, exclude specific models, or respect data, privacy, and compliance constraints, without retraining.
More speculatively, treating orchestration as a first-class scaling axis may have economic and geopolitical consequences, distributing the benefits of frontier AI more broadly across organizations and regions rather than concentrating them in those able to train the largest models.
We offer Fugu as a first step in this direction, and hope it encourages further research into collective intelligence as a path toward the next frontier of AI capabilities.

%% file: sections/02_background.tex
\textbf{LLM Collective Intelligence}. 
Collective intelligence studies how groups of relatively limited individuals can, through interaction, produce behavior that exceeds the capability of any single member \citep{ha2022collective,jiang2025comprehensive}.
This perspective has recently become increasingly relevant for LLMs, as frontier models from different providers exhibit complementary strengths and weaknesses.
A growing body of work therefore studies how to coordinate groups of LLM agents through communication protocols, voting mechanisms, routing policies, and learned collaboration structures \citep{du2023improving, wang2025mixtureofagents, dang2025multi, guha2024smoothie, zhuge2024gptswarm, yue2025masrouter, chen2024routerdc}.
Recent multi-agent LLM systems often rely on hand-designed collaboration patterns: for example, \citet{du2023improving} and \citet{wang2025mixtureofagents} orchestrate multiple agents across successive discussion or reasoning rounds, using fixed interaction structures to improve final answers.
Other works move toward learned or adaptive coordination.
\citet{guha2024smoothie} and \citet{yue2025masrouter} learn representations that map queries to suitable agents or topologies, \citet{zhuge2024gptswarm} treats collaboration as a learnable graph, and \citet{chen2024routerdc} learns a router that directs each query to a single best-matched agent.

Sakana Fugu follows this line of work in treating intelligence as an emergent property of coordinated model collectives.
However, our goal differs from that of many prior multi-agent systems in two respects.
First, rather than exposing a multi-agent workflow that users must design, tune, or operate, Sakana Fugu presents the collective as a single model interface.
Second, rather than relying solely on fixed communication patterns or single-step routing, Fugu models are themselves trained orchestrators that can adaptively decide how to use an agent pool for each query.
In this sense, Sakana Fugu aims to make collective intelligence not only a research paradigm for studying LLM collaboration, but also a practical model interface for accessing complementary model capabilities. 

\textbf{Agentic Scaffolds}.
To elevate LLMs from autoregressive chatbots into action-taking agents, a large body of work has developed agentic scaffolds that elicit reasoning, planning, reflection, and tool use from an underlying model.
A first line of work shapes the model's reasoning through prompting and decoding strategies, including chain-of-thought prompting \citep{wei2022chain}, self-consistency \citep{wang2022self}, and structured search over reasoning paths such as Tree of Thoughts \citep{yao2023tree}.
A second line interleaves reasoning with acting, allowing the model to plan, call tools, and incorporate observations: ReAct couples reasoning traces with actions \citep{yao2022react}, while Toolformer and subsequent work equip models with external tools and function calling \citep{schick2024toolformer, qu2025tool}.
A third line adds feedback and iteration, where models critique and revise their own outputs through self-reflection \citep{madaan2023self, shinn2023reflexion} or by incorporating signals from an external environment.
Together with growing support for long-horizon memory \citep{zhang2025survey}, these components transform a static model into a capable agent.

These ideas have also been consolidated into production harnesses that wrap frontier models in rich interactive environments.
In software engineering, for example, systems such as Claude Code and Codex manage repository context, tool use, code execution, editing loops, and environment feedback, often contributing substantially to end-to-end performance beyond the raw model call itself \citep{openai_codex_cli,anthropic_claude_code}.
These systems demonstrate that the practical capability of an LLM depends critically on the scaffold through which it interacts with the task environment.
However, such harnesses are typically designed for a specific class of tasks and expose the scaffold as an external system around the model.
Sakana Fugu takes a different approach, in which the scaffold is generated by a trained orchestrator at inference time. 
Rather than applying a fixed interaction pattern or relying on a task-specific harness, Fugu dynamically decides how to reason about each request, which frontier agents to involve, how they should communicate, and how their outputs should be synthesized into a final answer.

\textbf{Model Merging}.
Another line of work studies how to combine the capabilities of multiple models through model merging.
At the parameter level, early approaches use static recipes such as weight averaging, model soups, or task-balanced interpolation to integrate capabilities across domains with minimal additional computation \citep{goddard-etal-2024-arcees}.
More recent work introduces optimization-based merging, for example evolutionary search over ``merging recipes'' showing that learned strategies can outperform hand-designed ones and yield stronger generalization \citep{akiba2025evolutionary}.
Other methods improve the reliability of parameter-space fusion by resolving task conflicts, sparsifying parameter updates, or preserving directions important to individual models \citep{yadav2023ties, yu2024language}.
However, because parameter-space merging operates directly on weights, it requires access to model parameters and typically assumes architectural or representational compatibility.
This confines its applicability largely to open-source checkpoints and makes it unsuitable for composing the closed-source frontier models that currently define much of the state of the art.

A complementary family of methods composes models in the data-flow or representation space.
Instead of averaging all weights, these approaches stitch layers, mix blocks, route hidden states, or construct hybrid architectures from existing components \citep{bansal2021revisiting,akiba2025evolutionary}.
Such methods relax some constraints of full parameter merging and can combine capabilities at a more modular level, but they still generally require internal access to model activations, layers, or architectures.
They therefore cannot be generally applied to heterogeneous API-only systems, where models may differ in architecture, provider, context interface, latency, cost, and availability.

Sakana Fugu can be viewed as a macro-level analogue of model merging.
Rather than merging weights or stitching layers, it composes model capabilities at the behavioral level, treating frontier models as black-box agents and learning how to route, coordinate, verify, and synthesize their outputs.
In this sense, Fugu performs a form of functional model composition wherein it aims to preserve and amplify the complementary specializations of different models without requiring parameter access or architectural compatibility.
This macro-level perspective allows Fugu to incorporate closed-source frontier models, heterogeneous providers, and user-specific constraints, while still pursuing the central goal of model merging so that it combines multiple specialized models into a stronger collective system.

%% file: sections/03_sakana_fugu.tex
Sakana Fugu is a family of learned orchestrators that expose a multi-agent system through a single model interface.
Given a user query, a Fugu model constructs an agentic scaffold over a pool of frontier LLM workers, deciding which workers to involve, what instructions or roles to assign, how intermediate outputs should be combined or verified, and when to synthesize the final answer.
The user interacts with Fugu as if calling a single model, while internally the system can route, delegate, and coordinate across multiple specialized agents.
We release two variants targeting different points on the quality-latency frontier.
\textsc{\textbf{Fugu}} balances strong performance with low latency, making it suitable for everyday interactive use and configurable deployment constraints.
\textsc{\textbf{Fugu-Ultra}} prioritizes answer quality, using deeper orchestration over a larger worker pool at the cost of additional latency.

%% file: sections/03_fugumini.tex
\subsection{\textsc{Fugu}: Balancing Performance and Latency}
\label{sec:fugumini}

\textsc{Fugu} is the latency-aware variant of Sakana Fugu, designed for interactive and everyday workloads where response time matters alongside answer quality.
It builds on Trinity \citep{xu2025trinity}, but scales and adapts the learned-orchestration idea to a production setting in which the orchestrator must make fast, reliable routing decisions under reasonable latency and must be optimized not only for single-step performance but also for multi-turn end-to-end interactive tasks common in real-world use.

\subsubsection{Parametrization}
\label{sec:fugu-parametrization}

\begin{figure}[ht]
    \centering
    \includegraphics[width=0.9\textwidth]{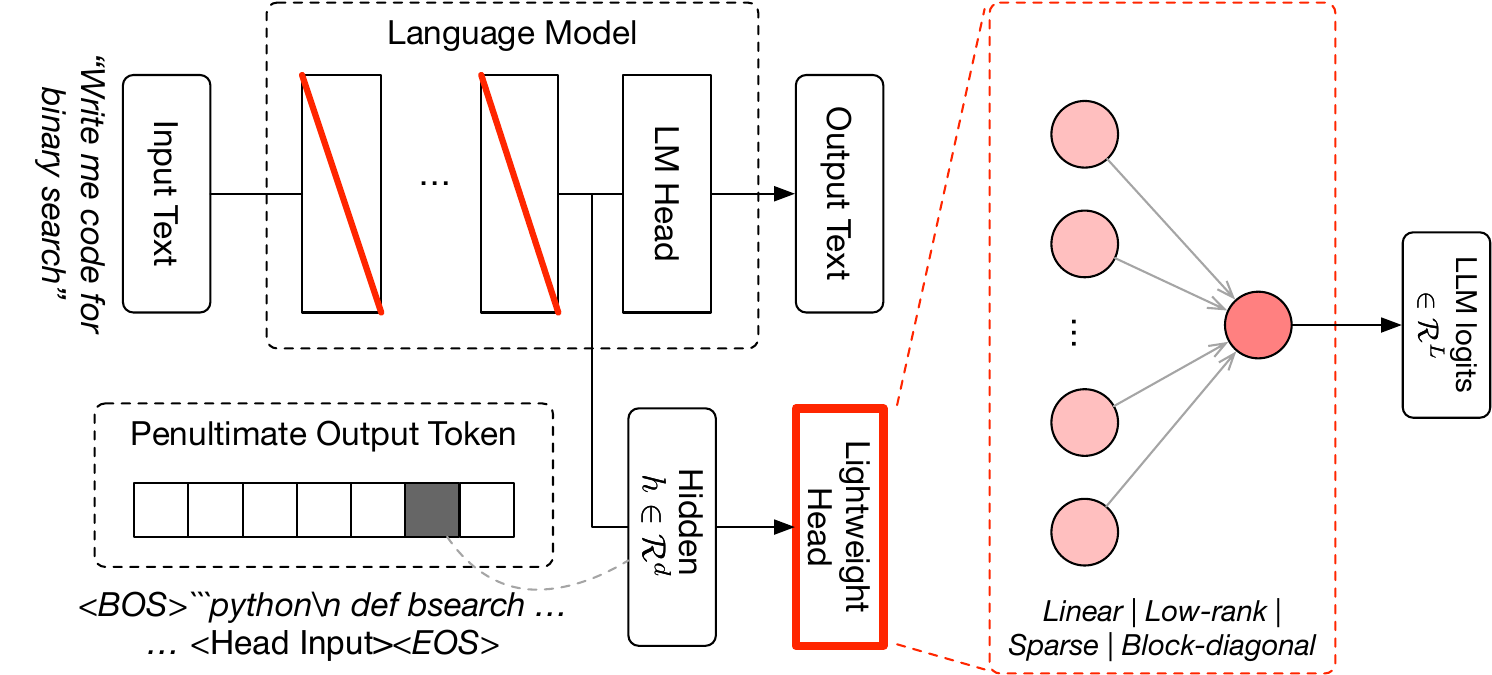}
    \caption{\textsc{Fugu} Parametrization. A lightweight selection head operates in parallel to the base model's LM head. It takes a hidden state $h$ from the orchestrator backbone as input and outputs $L$ logits, one for each worker model in the pool. Unlike the Trinity coordinator, \textsc{Fugu} does not assign roles. The selected model is always invoked as a worker, which reduces the coordination space and lowers orchestration latency. We also fine-tune the singular-value scales of selected parameter matrices in the LM's layers, indicated by the red diagonal lines. In the figure, the hidden state at the position marked by ``$<$Head Input$>$'' is the input to the lightweight head. The semantic correspondence between the decoded message ``$<$BOS$>$ ...'' and the hidden state is shown only for illustration, since the lightweight head operates on the internal hidden state at that position, not on the final decoded text.}
    \label{fig:parametrization}
\end{figure}

\textsc{Fugu}'s orchestrator is designed as a fast decision module over a pool of frontier worker models. Fugu uses a pre-trained language model as its backbone and coordinates the worker pool based on its own hidden states. We provide a high-level illustration of this process in Figure~\ref{fig:parametrization}.

Concretely, given a pool of $L$ worker models, we attach a lightweight prediction head after the final hidden layer of the orchestrator backbone.
For a hidden state $h \in \mathbb{R}^d$, the head outputs $L$ logits that score which worker model should be selected for the input.
Unlike Trinity, which additionally assigns one of several roles to the selected model, \textsc{Fugu} always dispatches the query to the selected model as a worker.
Removing role assignment narrows the coordination space to model selection alone, which keeps the orchestration decision simple and minimizes the latency overhead of the orchestrator.

To improve the representation used by this lightweight head without full fine-tuning, we adapt a small subset of the backbone parameters using singular-value fine-tuning, following recent work on efficient adaptation \citep{sun2025transformersquared}.
For selected weight matrices in the orchestrator backbone, we decompose the matrix and train only the singular-value scales, keeping the orthogonal components fixed.
Together with the lightweight prediction head, this yields an extremely small trainable parameter set while still allowing the orchestrator representation to effectively adapt to the routing problem.

A key design choice is that \textsc{Fugu} uses the orchestrator's logits rather than its generated text.
Since prompting and task execution are delegated to the selected frontier worker model, the orchestrator only needs to produce a worker-selection decision.
This makes inference substantially cheaper: the system can compute a hidden state at an early token position, apply the selection head, and immediately dispatch the query to the selected worker, without the expensive autoregressive decoding process. 
This decision-only parametrization is central to \textsc{Fugu}'s latency profile and also makes evolutionary optimization practical, as described in the following subsection.

\subsubsection{Supervised Fine-tuning on Single-Step Tasks}
\label{sec:fugu-sft}

We train \textsc{Fugu} in two stages, starting with large-scale supervised fine-tuning (SFT).
For this stage, we assemble a large collection of single-step tasks spanning coding, mathematics, reasoning, language understanding, and numerous agentic scenarios.
Each question $q_i$ in this collection $\mathcal{D}$ is verifiable and has a ground-truth solution $s_i$.

To construct training labels, we run every worker model $\mathcal{M}_j$ where $j =1 \cdots K$ in the pool on $q_i$ for $n$ repetitions and measure each model's performance by comparing with $s_i$, yielding for each model a set of candidate solutions $\{o^{\mathcal{M}_j}_1, o^{\mathcal{M}_j}_2, ..., o^{\mathcal{M}_j}_n \}$ and a corresponding reward set $\{r^{\mathcal{M}_j}_1, r^{\mathcal{M}_j}_2, ..., r^{\mathcal{M}_j}_n\}$.
These rewards induce a ranking over the worker pool, from the model best suited to the problem to the least suited.
We summarize for each model $j$ the performance on $q_i$ by the average of its rewards,
$
\bar{r}_{i,j} \;=\; \frac{1}{n}\sum_{k=1}^{n} r^{\mathcal{M}_j}_k,
$
and collect these scores into a vector $\mathbf{s}_i = (\bar{r}_{i,1}, \dots, \bar{r}_{i,K})$.

Rather than discarding the reward magnitudes by supervising on the hard ranking alone, we convert the scores into a soft target distribution over workers via a softmax transform,
\begin{equation}
    p_i(j) \;=\; \frac{\exp(\bar{r}_{i,j}/\tau)}{\sum_{j'=1}^{K} \exp(\bar{r}_{i,j'}/\tau)},
\end{equation}
with temperature $\tau$.
We use this distribution as supervision, training both the lightweight selection head and the singular-value scales in the orchestrator backbone to minimize:

\begin{equation}
\mathcal{L}_{\mathrm{SFT}}(\theta)
=
\frac{1}{|\mathcal{D}|}
\sum_{i=1}^{|\mathcal{D}|}
\mathbb{D}_{KL}\!\left(
    p_i(\cdot)
    \,\|\, 
    \pi_{\theta}(\cdot \mid q_i)
\right),    
\end{equation}

Learning from a soft performance distribution, rather than classifying a single best label, gives the orchestrator a richer training signal and makes its selection more robust when several workers are similarly capable.
Because supervision is derived directly from measured worker performance and does not require generation from the orchestrator itself, this stage provides an efficient and stable initialization for the subsequent evolutionary optimization stage.

\subsubsection{Applying Evolutionary Strategies on End-to-end Tasks}
\label{sec:fugu-es}

After supervised fine-tuning on single-step tasks, we further optimize \textsc{Fugu} with evolutionary strategies on end-to-end tasks. 
While single-step tasks provide clean correctness signals and broad coverage across domains, they do not fully capture how models are used in real interactive systems.
We therefore collect real-world multi-turn trajectories from different coding-assistant environments like Claude Code, Codex, and OpenCode, and construct end-to-end tasks that involve repository context, iterative editing, tool calls, execution feedback, and final task completion.
This stage expands the training distribution from static questions to agentic workflows that better reflect production usage.
Formally, for an end-to-end task $q_i$ drawn from a collection $\mathcal{E}$, the orchestrator interacts with the harness over a sequence of turns.
Let $s_t$ denote the interaction state at turn $t$, i.e., the task together with the transcript of prior turns, tool calls, and execution feedback.
At each turn, \textsc{Fugu} computes a hidden state $h(s_t)$ and selects a worker $a_t \sim \pi_{\theta}(\cdot \mid s_t)$ with $\pi_{\theta}(a \mid s) \propto \exp\!\big(f_{\theta}(h(s))_a\big)$, yielding a trajectory $\tau = (s_0, a_0, s_1, a_1, \dots, s_T)$ of horizon $T \le B$, where $B$ is a fixed turn budget.
A terminal reward $R(\tau) \in \{0, 1\}$ records whether the task is ultimately completed.

These end-to-end trajectories also reveal differences between agents that are not visible from performance scores alone.
Some models may be strong at high-level reasoning or proposing implementation plans, but less reliable when operating tools, editing files, or reacting to environment feedback.
Others may be less impressive on standalone reasoning benchmarks but more robust inside an interactive coding harness.
Training on these trajectories therefore helps \textsc{Fugu} learn a more practical notion of worker capability, one that accounts not only for answer quality in isolation, but also for how well each worker performs when embedded in a scaffold with tools and feedback.

For this stage, we use sep-CMA-ES, following the optimization approach of Trinity.
Concretely, we directly maximize the expected terminal reward of the orchestrator,
\begin{equation}
    J(\theta) \;:=\; \mathbb{E}_{\tau \sim \pi_{\theta}}\!\big[R(\tau)\big],
\end{equation}
over end-to-end task outcomes.
sep-CMA-ES maintains a parent parameter vector $\theta_t$, a step size $\sigma_t$, and a diagonal covariance $D_t$, and at each iteration samples a population of $\lambda$ candidates,
\begin{equation}
    \theta^{(k)} \;=\; \theta_t + \sigma_t\, D_t\, z^{(k)},
    \qquad z^{(k)} \sim \mathcal{N}(0, I),
    \quad k = 1, \dots, \lambda.
\end{equation}
Each candidate's fitness $J(\theta^{(k)})$ is estimated by averaging the terminal reward over replicated end-to-end runs, and the top-$\mu$ candidates are recombined via fitness-weighted averaging to form the next parent,
\begin{equation}
    \theta_{t+1} \;=\; \theta_t + \sigma_t\, D_t \sum_{j=1}^{\mu} w_j\, z_{j:\lambda},
\end{equation}
where $z_{j:\lambda}$ is the perturbation of the $j$-th best candidate under the fitness ranking and $\{w_j\}_{j=1}^{\mu}$ are the recombination weights.
Evolutionary optimization is well suited here because the SFT stage already places the orchestrator parameters in a strong region of the search space, allowing the evolutionary search to refine routing behavior at a finer granularity \citep{risi2026neuroevolution}.
It also lets us optimize end-to-end task outcomes directly, including sparse or noisy success signals, without constructing reliable ranking labels for complex multi-turn trajectories.
Empirically, we find this stage to be substantially more stable than supervised fine-tuning on the additional end-to-end tasks and better suited to settings where correctness depends on the full interaction between the worker model, the harness, tools, and environment feedback.

%% file: sections/03_fuguultra.tex
\subsection{\textsc{Fugu-Ultra}: Prioritizing Performance}
\label{sec:fuguultra}

\textsc{Fugu-Ultra} prioritizes unlocking the maximum capability from an LLM agent team and is designed for the most complex user workloads, where absolute answer quality takes precedence. 
\textsc{Fugu-Ultra} builds on the Conductor \citep{nielsen2025learning}, adding novel extensions to accommodate long-horizon function-calling and multi-agent workflows through adaptive agent memory.

\subsubsection{Conducting an Orchestra of Models}\label{sec: conductor}

We begin with a brief overview of the Conductor framework powering \textsc{Fugu-Ultra} before detailing the extensions to accommodate multi-agent function calling and shared memory. 
The Conductor framework leverages reinforcement learning for training a language model to prompt-engineer and coordinate a set of powerful LLM agents. 
The Conductor outputs full \textit{agentic workflows} as natural language that divide an input task, allocate arbitrary subtasks, and define targeted communication strategies to best make use of the agents' complementary capabilities.

\textbf{The Conductor task.} The Conductor's objective is to solve tasks \textit{indirectly} by designing \textit{agentic workflows} specific to any input question $q_i$. Each agentic workflow is defined as a sequence of \textit{workflow steps} 
whose final output is returned as the actual Conductor response $o_i$. 
Each step specifies a string with a natural-language \textit{subtask}, an integer id corresponding to the \textit{assigned worker agent} responsible for performing that subtask, and an \textit{access list} indexing which subtask solutions from the previous steps to include in the worker's context.
This design lets the Conductor freely craft tailored subtasks and communication strategies across its workers, allowing the specification of agentic workflows ranging from simple best-of-N and sequential chain-like topologies to arbitrary parallelizable tree-structured approaches, harnessing the individual strengths and synergies of its highly-specialized agents.

\textbf{Workflow execution and learning dynamics.} 
Each agentic workflow the Conductor outputs is executed by prompting the specified worker agent with their assigned subtask. 
In each workflow step, the worker's context includes the sequence of previous subtasks and corresponding responses defined in the access list. 
Analogously to the traditional RL framework, the reward $r_i$ for each response from the Conductor model is determined by two progressive conditions:
\begin{enumerate}
    \item The Conductor format condition, setting $r_i$ to 0 for responses from which the lists of subtasks, worker agents, and access lists cannot be parsed.
    \item The Conductor correctness condition, setting $r_i$ to 1 if the final output from executing a well-formatted agentic workflow $o_i$ matches the solution $s_i$ and to $0.5$ otherwise.
\end{enumerate}
We trained \textsc{Fugu-Ultra} with GRPO \citep{shao2024deepseekmath}:

\begin{equation}
J(\theta)=
\mathbb{E}_{q\sim D,\, \{o\}^G_1\sim\pi_\theta(\cdot\mid q)}
\left[
\frac{1}{G}\sum_{i=1}^G
\Big(
\min\!\big(
r_i A_i,\;
\mathrm{clip}(r_i,\,1-\epsilon,\,1+\epsilon)\,A_i
\big)
-\beta\,\mathbb{D}_{\mathrm{KL}}(\pi_\theta\,\|\,\pi_{\text{ref}})
\Big)
\right],
\label{eq:grpo}
\end{equation}

using the grouped completions to compute a Monte-Carlo \textit{advantage function}~\citep{sem_advantage_fn}:
\begin{equation}
A_i=\frac{r_i-\mathrm{mean}(\{r_1,\dots,r_G\})}{\mathrm{std}(\{r_1,\dots,r_G\})},
\label{eq:grpo_adv}
\end{equation}

where $r_i$ is our specific Conductor reward. The Conductor framework also allows specifying the orchestrator itself as a worker agent, further extending the range and complexity of possible coordination topologies.

Training \textsc{Fugu-Ultra} with this recipe, we observe the emergence of problem decompositions and prompt-engineered subtasks that leverage the differing strengths and skills of each agent in the team, alongside communication topologies that leverage both independent work and agent-to-agent collaboration according to the task requirements. 

\subsubsection{Function Calling Agentic Workflows}\label{sec: function call agentic}

Function calling within a multi-agent workflow poses a unique additional challenge in terms of orchestration memory. 
For a generic, single-agent system, the function call loop \citep{anthropic_tooluse, openai_functioncalling, mcp_intro} requires no additional persistent memory on the part of the agent, since the message transcript carries the full context and there is only one possible recipient to receive that context. 

In our \textsc{Fugu-Ultra} system, however, any agent may make a function call at any time. 
Thus, to honor the function call loop and enable any agent to interact freely with the user environment, the system must therefore retain \textit{which} agent emitted every call, along with where that agent sits in the overall Conductor workflow, such that each agent's function call loop is correctly routed to the corresponding agent and the inter-agent communication topology is maintained. 
This therefore requires our orchestrator to track the Conductor's \textit{workflow state}  containing the selected models, communication topology, and assigned subtasks for the corresponding user query throughout a user-agent interaction. 

\paragraph{Intra-workflow agent isolation.} In order to fully leverage the differing specialties of the agent team, we isolate each agent's function calling trajectory from one another. 
This intra-workflow isolation is necessary to prevent \textit{orchestration collapse}, whereby the first agent to interact with the environment sets the trajectory for all future agents, leading to redundant contributions from future agents as they are steered to follow the path initialized by the first agent. 
That is, an agent observes the actions and outputs of another agent \textit{only through the access list}, as proscribed by \textsc{Fugu-Ultra}, otherwise observing a transcript history preserving only their own actions. 
This prevents conditioning subsequent agents' solution trajectories on the work done by previous agents, allowing each agent full freedom to determine its own solution path as instructed by the subtask assigned to it by our model. 

\paragraph{Persistent shared memory.} While intra-workflow isolation is necessary to prevent orchestration collapse, complete isolation from all function calling over a multi-turn conversation history is likewise suboptimal. 
This is because agents must retain some memory of their interactions with the environment in order to not make redundant, repeated tool calls to rediscover the same artifacts and accrue the necessary background context to solve the required task. 
To resolve this tension, we permit inter-workflow shared memory across agents, which allows agents to observe tool calling from previous workflows. 
Thus, we grant agents full memory over the ongoing state of a multi-turn conversation, which provides the required background context to each agent, while isolating the agents from one another within the current workflow, but for the context determined necessary by \textsc{Fugu-Ultra}'s designated communication topology described by the access list.

\subsubsection{Training Setup}\label{sec: training}

To train \textsc{Fugu-Ultra}, we scale the Conductor's reinforcement learning approach starting from a pre-trained checkpoint of a regular language model. Ultra is instructed to design agentic workflows of up to 5 steps using a diverse pool of frontier LLMs that includes Gemini-3.1-Pro \citep{deepmind2026gemini31pro}, Claude-Opus-4.8 \citep{anthropic2026claudeopus48}, and GPT-5.5 \citep{openai2026gpt55}. For multi-turn tasks, any agent is permitted unlimited interaction with the user environment. We train using our Conductor reward detailed in section \ref{sec: conductor} with GRPO and without any KL divergence penalty. 

Our training dataset is a mixture of both publicly available data and expert-designed end-to-end environments simulating real agent-user interactions. These end-to-end tasks were also used in the development of our previous model (section \ref{sec:fugu-es}) and help expose different kinds of real-world specializations across agents, such as expertise in math, science, engineering, factual knowledge and recall, numerous tool-usage scenarios, conversational dialog, multi-turn context retention, and planning. Large-scale training on these tasks produces substantial gains in \textsc{Fugu-Ultra}, boosting its ability to discern the expertise and skills across the worker pool and how to optimally combine them. 

%% file: sections/04_results.tex
In this section, we report the capabilities of Sakana Fugu, as measured by performance on a diverse set of challenging benchmarks (section \ref{sec: benchmark}), as well as expert-designed tasks (section \ref{sec: expert-designed tasks}), which were hand-crafted to be representative of real and challenging use cases for Sakana Fugu. We additionally analyze the orchestration adaptivity of Sakana Fugu across domains (section \ref{sec: domain adaptivity}), before concluding with findings on optimal coordination strategies and agent topologies (section \ref{sec: optimal topologies}).

\subsection{Benchmark Performance}\label{sec: benchmark}

We first evaluate Sakana Fugu on a broad suite of standardized benchmarks designed to measure core agentic capabilities under reproducible conditions.
These benchmarks cover tasks requiring reasoning, tool use, coding, long-horizon problem solving, and domain-specific expertise, providing a quantitative view of Fugu's performance relative to frontier baselines.

\subsubsection{Evaluation setup}

We tested the capabilities of the Sakana Fugu models on contemporary, challenging benchmarks unseen by our models during training to evaluate their generalization capabilities. For agentic coding and software engineering, we use SWE Bench Pro \citep{deng2025swe} and Terminal Bench 2.1 \citep{merrill2026terminalbenchbenchmarkingagentshard}. To best expose the underlying capabilities of Fugu, we use minimal evaluation harnesses such as Mini-SWE-agent \citep{mini_swe_agent} and Terminus 2 \citep{harbor_terminus2}. To evaluate scientific knowledge, we used GPQA Diamond \citep{rein2024gpqa}, the set of diamond difficulty questions on natural science taken from the Graduate-level Google-proof Q\&A benchmark. For multidisciplinary reasoning, we used Humanity's Last Exam \citep{phan2025humanity}, including multimodal samples, and evaluated without tools. 
To evaluate competitive programming, we used the latest full v6 release of LiveCodeBench \citep{jain2024livecodebench} (problems taken from May 2023 to April 2025) and the latest publicly available and gradable split of LiveCodeBench Pro \citep{zheng2026livecodebench} (problems taken from quarter 2 of 2025). For additional coding evaluation with scientific application, we used SciCode \citep{tian2024scicode} with background provided. For multimodal graphical understanding, we used CharXiv Reasoning \citep{wang2024charxiv}, and for conversational dialog, we used $\tau^3$ Banking with GPT-5.2 as the user simulator. Finally, for long context retrieval and reasoning, we used needle-in-a-haystack MRCRv2 \citep{vodrahalli2024michelangelo} and long document information retrieval benchmark Long Context Reasoning \citep{artificialanalysis2025lcr}.
See Appendix \ref{app: eval details} for further details on benchmark evaluation configuration. 

Sakana Fugu models orchestrate a large pool of diverse models that includes state-of-the-art (SOTA) frontier LLMs such as Gemini-3.1-Pro \citep{deepmind2026gemini31pro}, Claude-Opus-4.8 \citep{anthropic2026claudeopus48}, and GPT-5.5 \citep{openai2026gpt55}. Therefore, to assess Fugu's ability to harness and amplify the skills of its individual workers, we evaluate Fugu against these same models, with the same maximum reasoning effort configured.

\begin{table}[ht]
\centering
\caption{Model Card. Fugu models, through intelligent orchestration of leading frontier models, harness and amplify their differing skillsets to achieve new SOTA capabilities. Best scores are in \textbf{bold} and second-best are \underline{underlined}. Baseline scores are provider-reported wherever available. Additional details in Appendix \ref{app: eval details}.}
\begin{tblr}{
  colspec      = {l c c c c c c},
  column{2}    = {rightsep=4pt},
  column{4}    = {leftsep=4pt},
  column{3}    = {leftsep=2pt, rightsep=2pt},
  cell{2-12}{2} = {bg=fugublue},
  cell{2-12}{4} = {bg=fugublue},
  row{1}       = {font=\bfseries},
}
\toprule
 & \textsc{Fugu-Ultra} & & \textsc{Fugu} & Claude Opus 4.8 & Gemini 3.1 & GPT-5.5 \\
\midrule
SWE Bench Pro          & \best{73.7}   & & 59.0          & \second{69.2} & 54.2 & 58.6  \\
Terminal Bench 2.1     & \best{82.1}   & & \second{80.2} & 74.6 & 70.3 & 78.2  \\
LiveCodeBench       & \best{93.2}   & & \second{92.9}   & 87.8 & 88.5 & 85.3  \\
LiveCodeBench Pro      & \best{90.8}   & & 87.8          & 84.8 & 82.9 & \second{88.4}  \\
Humanity's Last Exam   & \best{50.0}   & & 47.2          & \second{49.8} & 44.4 & 41.4  \\
CharXiv Reasoning      & \best{86.6}   & & \second{85.1} & 84.2 & 83.3 & 84.1  \\
GPQA Diamond                 & \best{95.5} & & \best{95.5}   & 92.0 & \second{94.3} & 93.6  \\
SciCode                & 58.7          & & \best{60.1}   & 53.5 & \second{58.9} & 56.1  \\
$\tau^3$ Banking       & \second{20.6} & & \best{21.7}   & \second{20.6} & 8.4 & \second{20.6} \\
Long Context Reasoning & 73.3          & & \best{74.7}   & 67.7 & 72.7 & \second{74.3}  \\
MRCRv2                 & \second{93.6} & & 86.6          & 87.9 & 84.9 & \best{94.8}  \\
\bottomrule
\end{tblr}
\label{tab:benchmarks}
\end{table}

\subsubsection{Agentic Coding}\label{sec: agentic coding}

\begin{figure}[ht]
    \centering
    \includegraphics[width=0.9\linewidth]{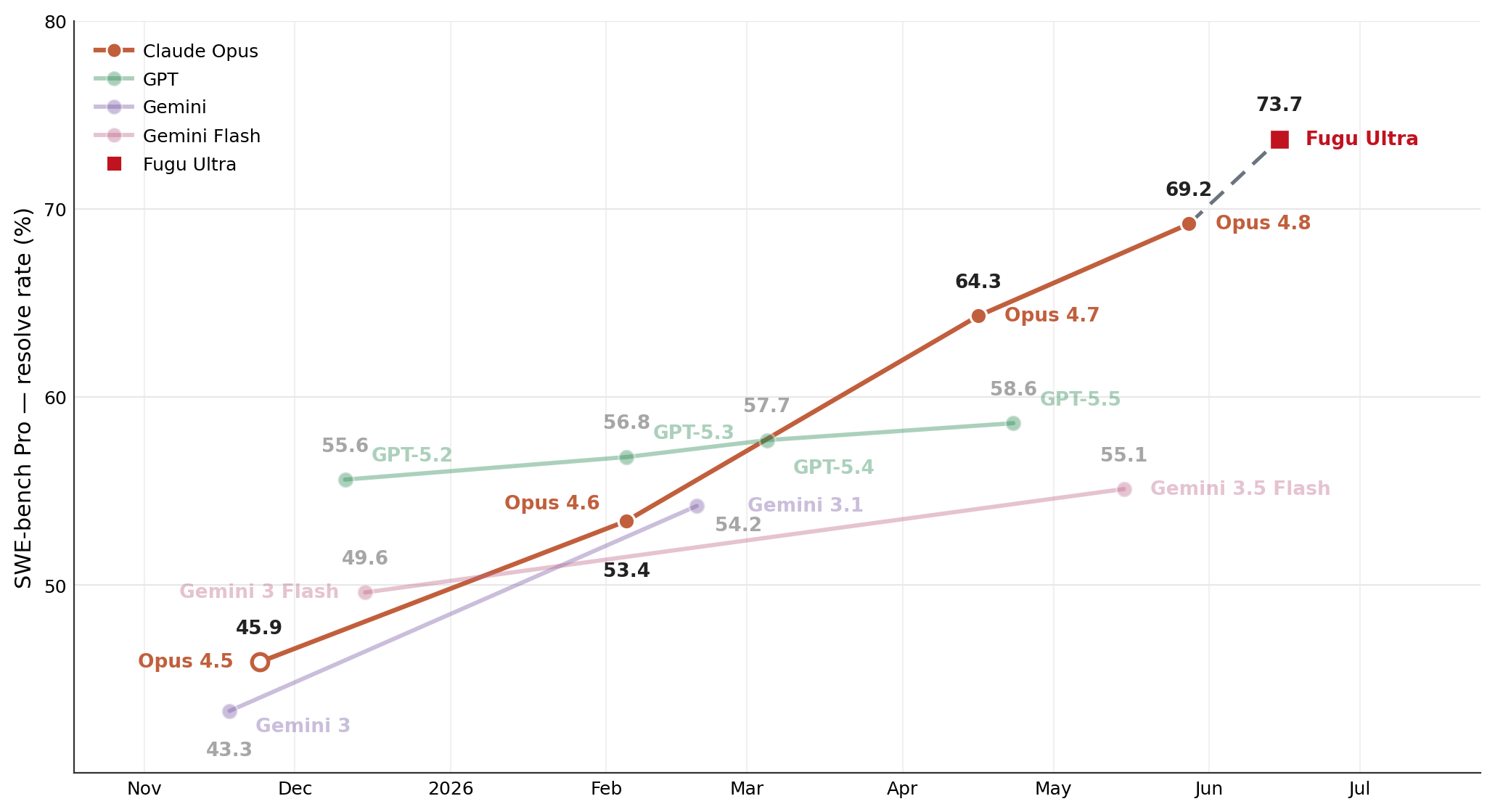}
    \caption{\textsc{Fugu-Ultra}'s performance gain in agentic coding is consistent with generational upgrades. Through intelligent orchestration of Opus-4.8, GPT-5.5, and Gemini-3.1, \textsc{Fugu-Ultra} accesses performance typically associated with the next iteration of model training.}
    \label{fig: swe generational}
\end{figure}

We find that \textsc{Fugu-Ultra} excels in agentic coding and software engineering, achieving state-of-the-art performance in both SWE Bench Pro and Terminal Bench 2.1, with a substantial leap forward in performance across both benchmarks. The gains in both benchmarks, both in the range of 5-6\% relative to the next best performer, are consistent with entire generational improvements across these frontier providers, as shown in Figure \ref{fig: swe generational}.

We additionally note that excelling across both benchmarks illustrates \textsc{Fugu-Ultra}'s fine-grained adaptivity, achieving and surpassing individual SOTA models in our agentic coding tasks, outperforming Claude-Opus in SWE Bench Pro and GPT in Terminal Bench. This demonstrates \textsc{Fugu-Ultra} successfully learns the granular capabilities of the agent team, including how they differ and how they can be optimally combined, to excel itself as an effective agent beyond the best of its individual workers. 

\textsc{Fugu}'s performance in Terminal Bench additionally reveals how per-step adaptivity can yield optimal performance. Despite \textsc{Fugu} only selecting one model for each input, we find that \textsc{Fugu} can also substantially outperform the state-of-the-art GPT-5.5 in this task. Analyzing the trajectories, we find \textsc{Fugu} alternates between GPT-5.5 and Claude-Opus-4.8 throughout the progression of the solution, calling in Claude-Opus-4.8 at particular, critical debugging points. Additional qualitative analysis on optimal trajectories is found in section \ref{sec: optimal topologies}.

\begin{figure}[ht]
    \centering
    \includegraphics[width=0.9\textwidth]{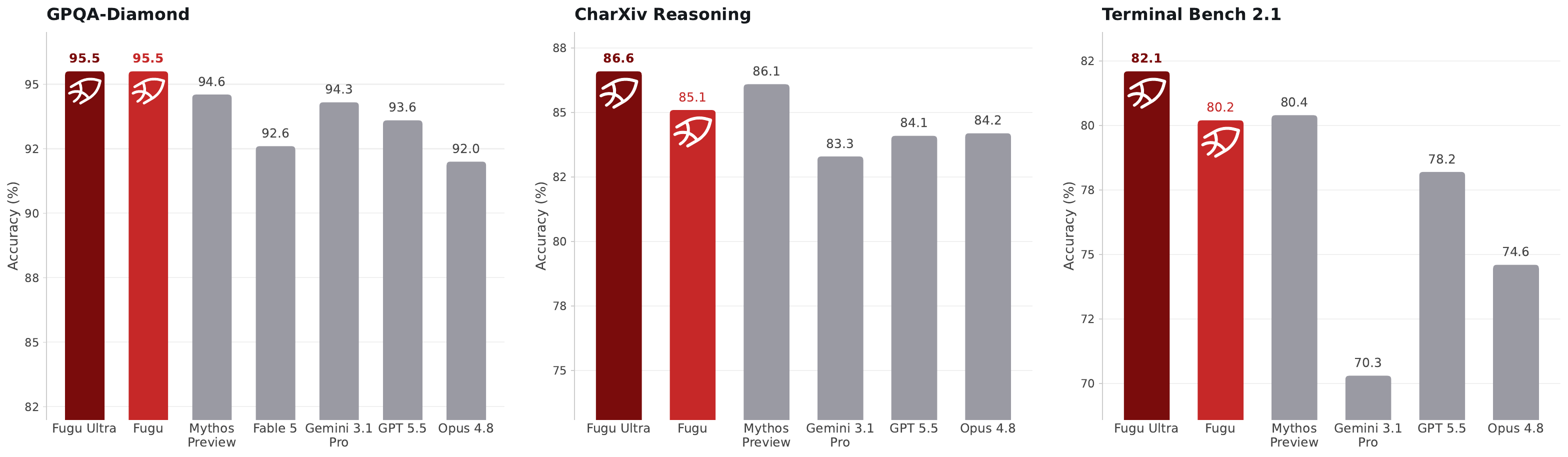}
    \caption{Fugu models exceed the capabilities of the Mythos Preview and Fable 5 model class purely through intelligent orchestration.}
    \label{fig:gpqad and charxiv}
\end{figure}

\begin{figure}[ht]
    \centering
    \includegraphics[width=0.95\linewidth]{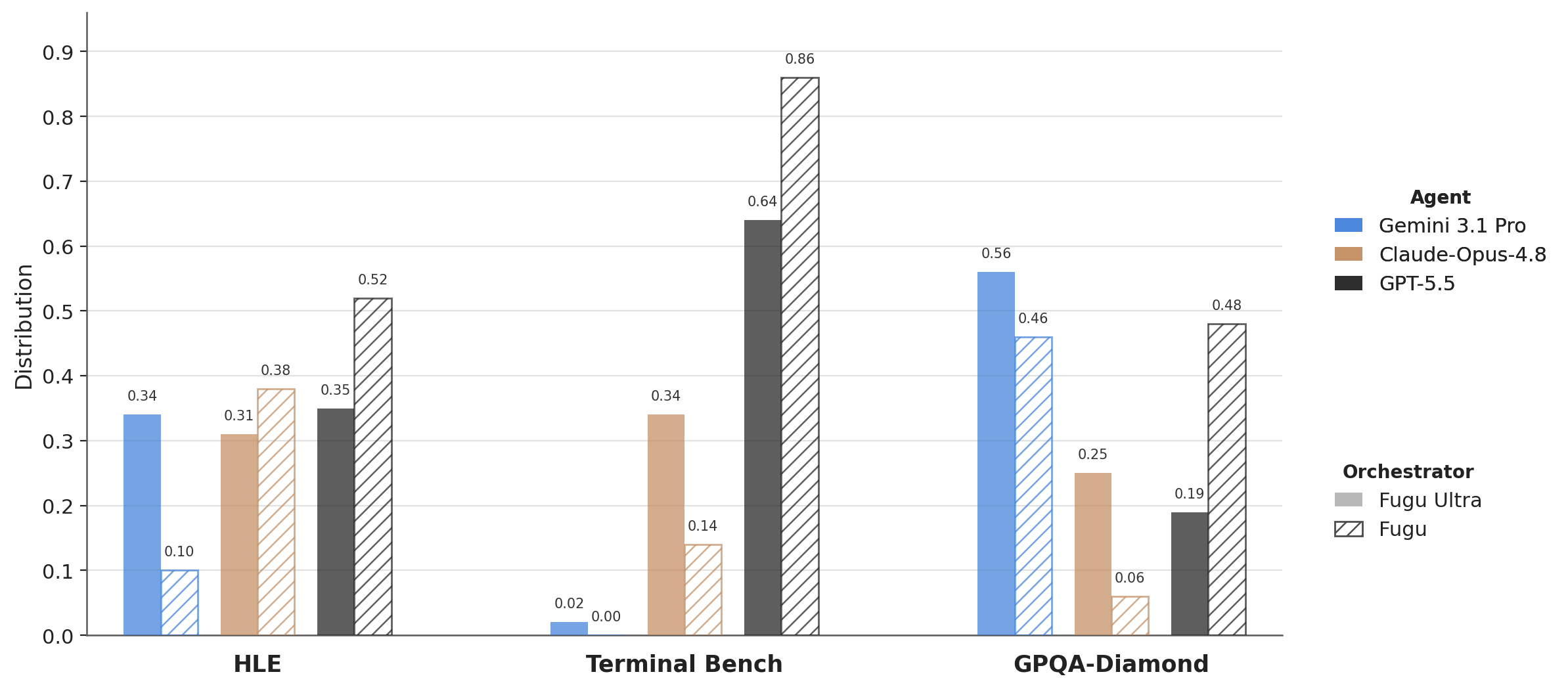}
    \caption{Per-task model distribution. Fugu models adapt their orchestration strategies across domains.}
    \label{fig: distribution}
\end{figure}

\subsubsection{Scientific Reasoning}\label{sec: multidisciplinary reasoning}

We additionally find that Sakana\textsc{Fugu} models excel in scientific reasoning, with both \textsc{Fugu} and \textsc{Fugu-Ultra} achieving SOTA performance in GPQA-Diamond. These results consolidate our models' ability to discern scientific expertise from their agent pool and indeed surpass the best-in-class performance of Gemini-3.1-Pro through selectively drawing on the complementary skillsets of additional models. A consistent pattern we observed was that our models recognized GPT's dominance in math and physics and applied GPT in targeted settings where mathematical computation was required. This fine-grained adaption to draw on GPT's mathematical expertise allowed our Fugu models to attain new levels of performance in scientific reasoning. 

Sakana Fugu models also exhibit such marked performance gains in scientific reasoning, even surpassing the Mythos Preview and Fable 5 \citep{anthropic2026claudefablemythos5} model class, which is not publicly available (see Figure \ref{fig:gpqad and charxiv}). Such a finding offers additional evidence of one of our central motivations for training Sakana Fugu, which is that intelligent orchestration ought to be an additional axis to scale performance without scaling training compute. Through targeted selection and application of existing frontier models, our models attain performance beyond both generational upgrades and even entire model classes.

\subsection{Domain Adaptivity}\label{sec: domain adaptivity}

We view adaptivity as a hallmark feature of an intelligent orchestrator. Throughout our evaluations, we find that Fugu models show consistent and varied adaptivity across their routing distributions, demonstrating the capability of our models to accurately learn the differing skills across the model team and then deploy these models according to those specializations.

Across domains, we observe deployed distributions agreeing with our SOTA capability priors. For example, in Figure \ref{fig: distribution}, we observe the agent distribution peaks on GPT-5.5 in Terminal Bench for both \textsc{Fugu} and \textsc{Fugu Ultra}, for which GPT-5.5 exhibits SOTA performance. Likewise, in GPQA-Diamond, where Gemini-3.1-Pro is the leading model, both Fugu models focus their orchestration around Gemini. In Humanity's Last Exam, which is by nature multidisciplinary, we observe ahighly balanced distribution over the three agents in \textsc{Fugu-Ultra}. Per-category, however, we observe that math questions are predominantly handled by GPT-5.5, again echoing widespread findings on GPT's expertise in math. Likewise, chemistry and biology questions are predominantly routed to Gemini, mirroring its SOTA scientific capabilities.

\subsection{Performance Beyond Benchmarks}
\label{sec: expert-designed tasks}

To complement aggregate benchmark scores, we evaluate Sakana Fugu on qualitative, end-to-end tasks that stress real agentic behavior such as long-horizon research, program synthesis, optimization, CAD generation, among others.
In these examples we compare Sakana Fugu models against three frontier baselines, namely Gemini 3.1 Pro (high), Opus 4.8 (max), and GPT 5.5 (xhigh).
We anonymize these baselines as Model A, Model B, and Model C in the descriptions, and we intentionally vary the mapping across examples to keep the discussion focused on behavioral differences rather than model identity or brand-specific expectations.
To ensure fair comparisons, we use identical settings for all the models in each experiment, so that any differences reflect model behavior rather than differences in tooling or orchestration infrastructure.
As an AI research lab, we show three research related examples in this section, and present more results in Appendix \ref{app:more_exp}.

\subsubsection{Autonomous LLM Training Optimization on AutoResearch}

We evaluate whether \textsc{Fugu-Ultra} can improve autonomous ML research workflows relative to strong single-model agents.
The task is drawn from AutoResearch~\citep{karpathy2026autoresearch}: an agent iteratively edits a small GPT training pipeline, executes each proposed change, and retains modifications that reduce validation bits-per-byte (BPB; lower is better).
This benchmark is well suited to our setting because success depends not on a single coding step, but on sustained exploration of optimizer settings, batching, architecture scale, and schedule choices over many sequential experiments.

All systems use the same AutoResearch scaffold, dataset, evaluation protocol, and per-experiment compute budget on a single H100 GPU.
Each agent is run for 123 autonomous experiments (${\sim}$14 hours total wall-clock time per seed).
We compare \textsc{Fugu-Ultra} against three frontier-model baselines, denoted Model~A, Model~B, and Model~C, each evaluated with maximum reasoning effort.
Each system is evaluated over three independent random seeds.
The only methodological difference beyond the shared scaffold is the agent backend: the baselines are single frontier models acting alone, whereas \textsc{Fugu-Ultra} orchestrates multiple strong models within a unified research loop.

\begin{figure}[h]
    \centering
    \includegraphics[width=0.8\linewidth]{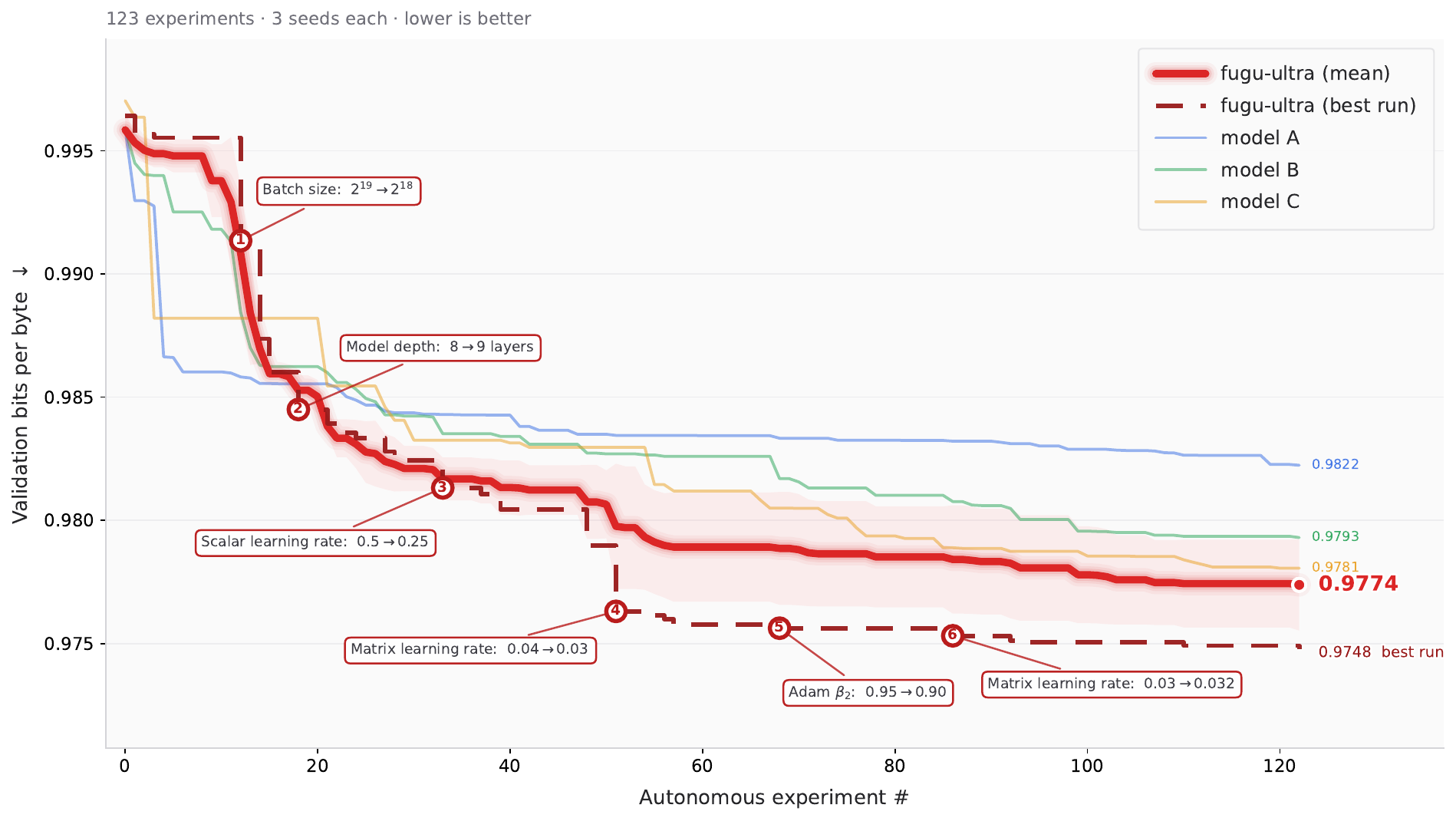}
    \caption{Validation bits-per-byte (BPB) as a function of the iteration of autonomous experiment on AutoResearch. \textsc{Fugu-Ultra} (red) is compared against Model~A (blue), Model~B (green), and Model~C (yellow), each averaged over three seeds. Lower is better.}
    \label{fig:autoresearch}
\end{figure}

Figure~\ref{fig:autoresearch} summarizes the optimization trajectories.
Solid lines report the mean best validation BPB across seeds (shaded bands: $\pm 1$ standard deviation); the dashed line shows the best single-seed run for \textsc{Fugu-Ultra} only.
Baseline trajectories are shown as mean $\pm$ std so the figure emphasizes cross-seed trends without cluttering the plot with four additional best-run curves.
Table~\ref{tab:autoresearch} reports the corresponding endpoint numbers for all systems, including each baseline's best single-seed result.

\begin{table}[t]
\centering
\caption{Final validation BPB after 123 autonomous experiments (lower is better).
Mean $\pm$ std is computed over three seeds; best run is the lowest BPB achieved by any single seed.}
\label{tab:autoresearch}
\begin{tabular}{lcc}
\toprule
Agent & Mean best val.\ BPB & Best single-seed run \\
\midrule
\textsc{Fugu-Ultra} & \textbf{0.9774 $\pm$ 0.0019} & \textbf{0.9748} \\
Model~C            & 0.9781 $\pm$ 0.0011          & 0.9766 \\
Model~B            & 0.9793 $\pm$ 0.0025          & 0.9758 \\
Model~A            & 0.9822 $\pm$ 0.0017          & 0.9799 \\
\bottomrule
\end{tabular}
\end{table}

At the end of the 123-experiment budget, \textsc{Fugu-Ultra} achieves the lowest mean validation BPB (\textbf{0.9774 $\pm$ 0.0019}), outperforming Model~C (\textbf{0.9781 $\pm$ 0.0011}), Model~B (\textbf{0.9793 $\pm$ 0.0025}), and Model~A (\textbf{0.9822 $\pm$ 0.0017}).
On best single-seed runs (Table~\ref{tab:autoresearch}), \textsc{Fugu-Ultra} reaches \textbf{0.9748} BPB, compared with 0.9766 (Model~C), 0.9758 (Model~B), and 0.9799 (Model~A), corresponding to absolute gains of 0.0018, 0.0010, and 0.0051~BPB, respectively.
These gains are modest in absolute magnitude, as expected for a highly optimized training pipeline, but they are consistent across both mean and best-run metrics and persist throughout the optimization process rather than emerging at a single checkpoint.

Two patterns are especially informative.
First, \textsc{Fugu-Ultra} is competitive early in the run and pulls ahead after mid-training, suggesting that orchestration is most valuable once the search space shifts from coarse configuration changes to finer optimizer and schedule tuning.
Second, although Model~B attains a strong best-seed result, its higher cross-seed variance in Table~\ref{tab:autoresearch} indicates less reliable optimization; \textsc{Fugu-Ultra} improves both peak performance and consistency.
Taken together, these results provide an initial demonstration that multi-model orchestration can outperform any individual frontier agent on an agentic training-optimization benchmark, supporting the broader claim that \textsc{Fugu-Ultra} is effective for autonomous ML research workflows.

\subsubsection{Classical Japanese Letter Reading Order}
Recovering the reading order of classical Japanese kana letters (kana-shōsoku) remains an open problem, with no established method and no public benchmark. To our knowledge no dataset for this task exists; we therefore construct our own, 25 pages hand-annotated by a domain expert, for this experiment. Unlike printed or even ordinary handwritten text, these letters are composed in the chirashigaki ("scattered writing") style: the characters are deliberately spread across the page at varying sizes and positions. Recovering the reading order is a task that even trained readers of classical Japanese find demanding. This is exactly the regime where data-driven learning does not apply, not because a learned model would be weak, but because the training data does not, and cannot readily, exist.

\begin{figure}[ht]
    \centering
    \includegraphics[width=0.75\linewidth]{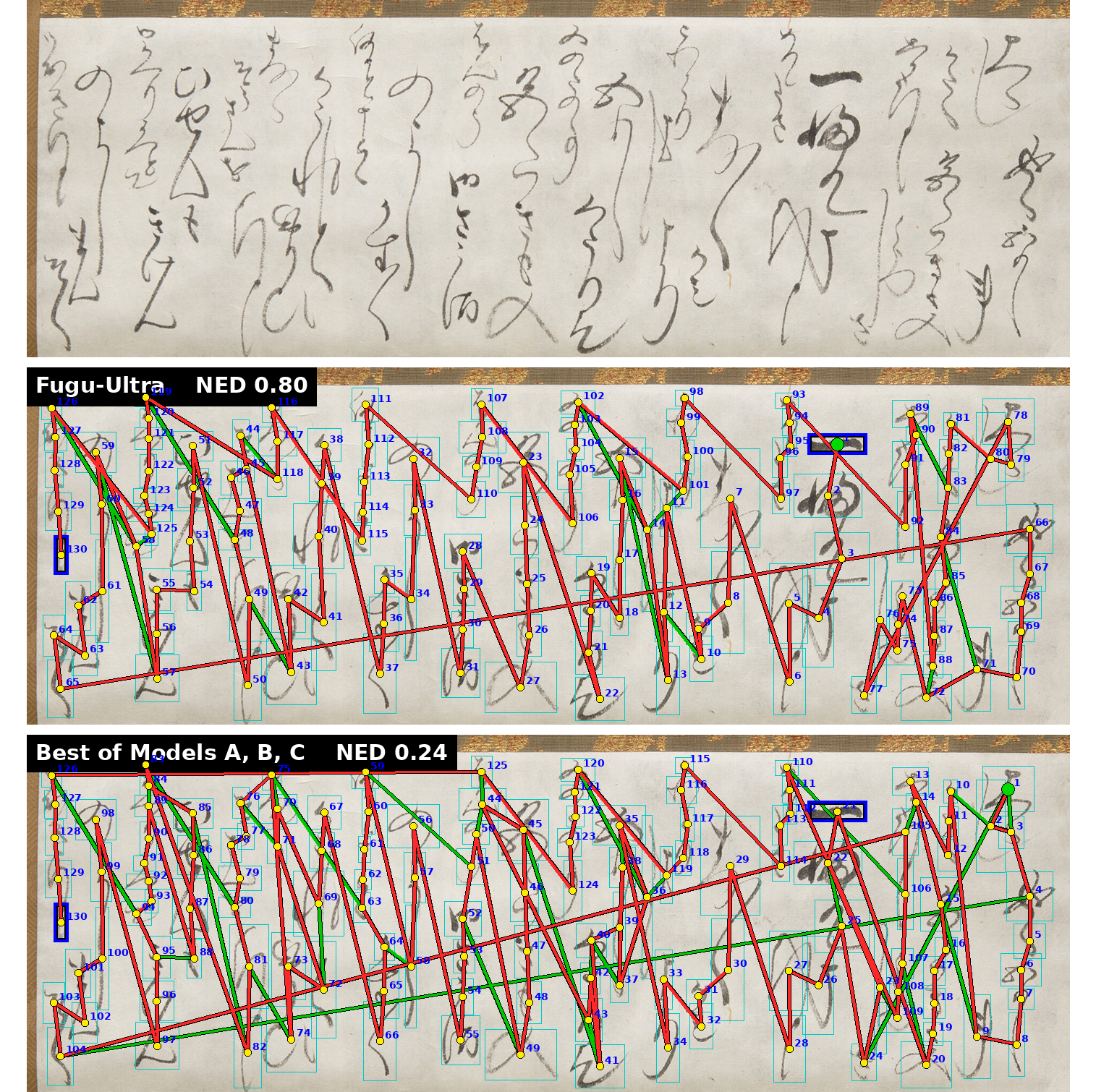}
    \caption{Reading-order recovery on one classical kana letter (kana-shōsoku). Top: the original letter, shown without annotation. Middle: \textsc{Fugu-Ultra} (NED 0.80 on this page). Bottom: a frontier baseline (NED 0.24). The predicted reading path (red) is overlaid on the expert ground truth (green); \textsc{Fugu-Ultra} tracks the expert's traversal of the scattered characters while the baseline does not. On this particular page both frontier models score about 0.24; the full-corpus numbers are in Table~\ref{tab:kana}. (Letter held by the Keio Institute of Oriental Classics.)}
    \label{fig:kanashosoku}
\end{figure}

\begin{table}[h]
\centering
\caption{Reading-order recovery on 25 expert-annotated kana letters; mean NED over all pages (higher is better; a seed heuristic scores $0.116$). Every model is improved by the same beam search, so the result tracks the model driving it: \textsc{Fugu-Ultra} is best, ahead of every frontier baseline, though the strongest frontier (Model~A) is competitive. A third frontier model did not complete the search at high effort. Models A and B are anonymized frontier baselines.}
\begin{tabular}{lc}
\hline
\textbf{Model} & \textbf{Mean NED} \\
\hline
\textbf{\textsc{Fugu-Ultra}} & \textbf{0.776} \\
Model A             & 0.642 \\
\textsc{Fugu}       & 0.473 \\
Model B             & 0.449 \\
Model C             & No completed run \\
\hline
Baseline & 0.116 \\
\hline
\end{tabular}
\label{tab:kana}
\end{table}

A domain expert can nonetheless state the rules they read by: characters are grouped by relative size, separated into distinct vertical blocks. The hard part is turning this tacit, qualitative rule set into a working algorithm. We test whether \textsc{Fugu} models can perform that translation directly. Rather than train a model, \textsc{Fugu} writes a reading-order predictor (a function over the detected character bounding boxes that returns their reading order) and improves it through test-time scaling. In this case we used beam search: at each iteration the model proposes new predictor variants, each is executed and scored, and the best are carried forward and mutated again over many rounds. The objective is normalized edit distance (NED) against the annotated reading order. 

The same beam search drives every model, and the outcome tracks the model driving it (Table \ref{tab:kana}). Across all 25 pages, \textsc{Fugu-Ultra} produces the best predictor, with a mean NED of 0.776, ahead of every frontier baseline. The strongest frontier baseline reaches 0.642, \textsc{Fugu} 0.473, and the weaker frontier baseline 0.449 (a seed heuristic scores 0.116). Figure \ref{fig:kanashosoku} shows one page on which \textsc{Fugu-Ultra} follows the expert's traversal across the scattered characters while the frontier predictors do not. With the search held fixed, \textsc{Fugu-Ultra} is the strongest engine, ahead of every frontier model, on a task where the model, not the search, sets the ceiling.

\subsubsection{CAD Generation}
In this experiment, we evaluate how well each model can generate CAD with a working mechanism from a text instruction. We chose a mechanical iris, the mechanism used in a camera aperture. This task requires more than a static circular part. The model must produce several blades, outer pins, a rotating ring, and a central opening whose geometry stays consistent as the iris opens and closes. Each model generated its design in Codex using the CAD skill~\citep{cadskills}, and we qualitatively evaluated the resulting STEP files. To check the dynamic behavior, we added a rule-based open-close animation driven by the blade pin positions.

\begin{figure}[h]
    \centering
    \includegraphics[width=\linewidth]{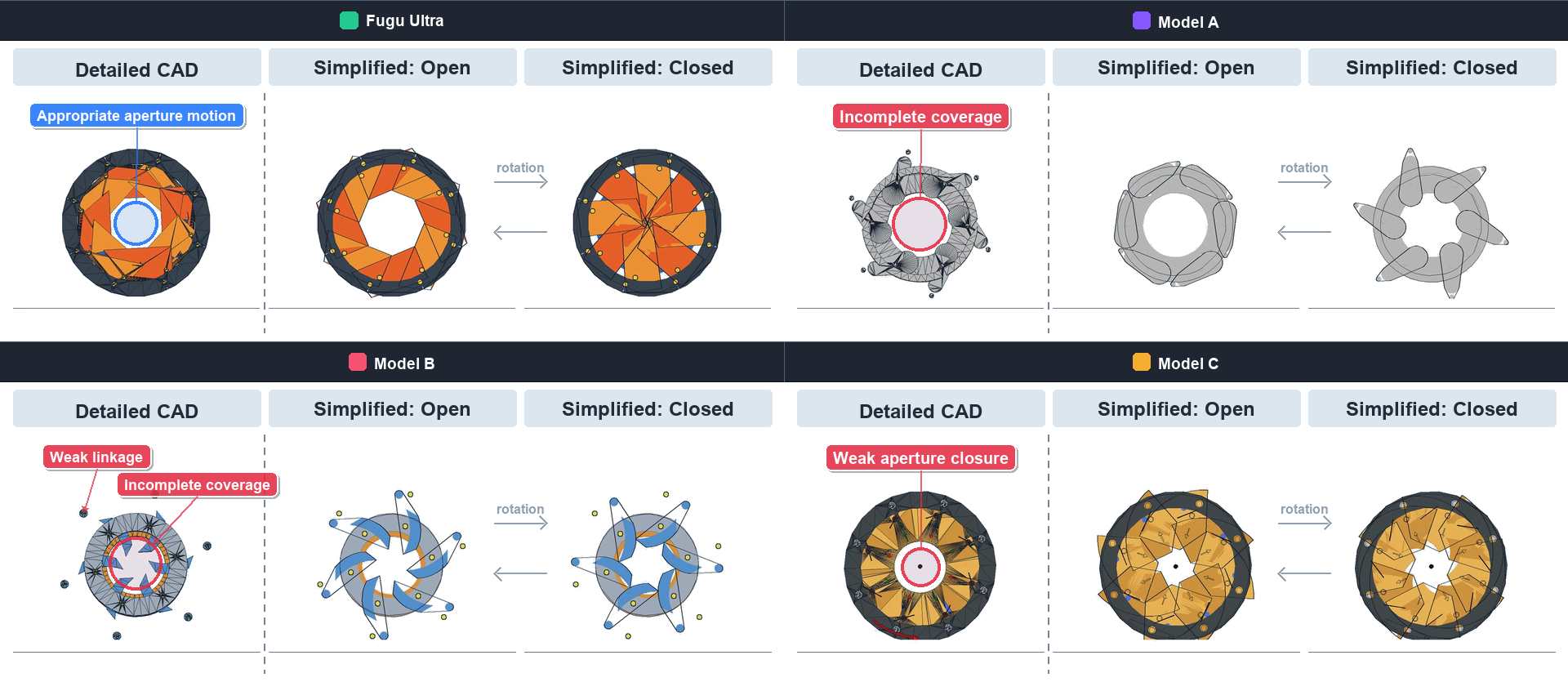}
    \caption{Mechanical iris generated in Codex with a CAD skill. We compare the CAD view (detailed) and the simplified view in the open and closed states. Blue marks the correct open-close region of \textsc{Fugu-Ultra}; red marks the incomplete coverage and weak links of the other models. \textsc{Fugu-Ultra} produces blades that rotate around their outer pins and a central opening that widens and narrows smoothly. In contrast, the other models fail to fully cover the center, form mechanically weak outer links, or do not close the opening enough.}
    \label{fig:iris_cad}
\end{figure}

Figure~\ref{fig:iris_cad} shows the CAD view and the simplified view in both the open and closed states. \textsc{Fugu-Ultra} generated a structure in which each blade rotates around its outer pin and smoothly widens or narrows the central opening. The other models were less complete as iris mechanisms: their blades did not fully cover the center, their outer links were mechanically weak, or the opening did not close far enough. We note that the results can vary with the prompt and across trials, but \textsc{Fugu-Ultra} consistently produced qualitatively strong results.

\subsection{Optimal Strategies and Topologies}\label{sec: optimal topologies}

We conclude with findings on performant orchestrative strategies, demonstrating granular application of agent expertise to achieve the high performance reported throughout this section. 

\paragraph{Debate and aggregation.} A core benefit of \textsc{Fugu-Ultra}'s dynamic workflows is that \textsc{Fugu-Ultra} can produce any coordination topology describable in natural language on a per-question basis. In knowledge-intensive domains, multi-round debate and tree topologies are powerful strategies to maximize the collective knowledge from an agent team. We find that such strategies indeed arise in domains requiring niche or specialized factual knowledge. For instance, in Humanity's Last Exam, when \textsc{Fugu-Ultra} was tasked to determine the outcome of a specific game state in \textit{Heroes of Might \& Magic III} -- a question testing both knowledge and application of game mechanics with precise trivia knowledge -- \textsc{Fugu-Ultra} produced a tree structure with Gemini-3.1-Pro at the head of the tree tasked to synthesize two independent attempts. Usage of Gemini as the aggregator made use of Gemini's SOTA performance in niche factual recall, however, further noteworthy, \textsc{Fugu-Ultra} also used a second instance of Gemini alongside GPT in the two leaf positions to attempt the problem independently. In this task, both leaf-agents produced partially correct answers, with Gemini incorrectly inferring a base defense statistic and GPT incorrectly applying a game-mechanical rule, however the Gemini-as-aggregator correctly identified the partial correctness in either answer and synthesized a fully correct answer. In a second example, when tasked to compute the Crowley–Nordström invariant of the Calabi–Yau link defined by a given weighted-homogeneous polynomial, \textsc{Fugu-Ultra}, leveraged the mathematical expertise of GPT at the root of the tree with both Gemini and Opus at leaf positions. Here, GPT-as-aggregator resolved a disagreement between Gemini and Opus over a single integer in the solution by rederiving the spectral numbers from the spectral generating function, locating the exact error and crux of the question's difficulty, and solving the question.

These simple examples are revealing in two ways. First, once again, we observe granular adaptivity on a per-question basis, with \textsc{Fugu-Ultra} utilizing tree-like topologies in both tasks while adapting its aggregator between Gemini for a trivia-heavy task and GPT for a math-heavy task, leveraging their individual expertise in the key aggregation role. Second, dynamic adaptation of an aggregator role is precisely the kind of adaptation unavailable to existing multi-agent systems \citep{wang2025mixtureofagents, openrouter2026fusion}, which necessitate a fixed model to \textit{always} act as a final synthesizer, regardless of whether that model is suitable for the task or not. Such systems are thereby bottlenecked by that rigidity, and typically struggle to surpass the performance of the aggregator for tasks outside of the aggregator's expertise. \textsc{Fugu-Ultra}'s adaptivity sidesteps this issue, and is key to unlocking the capabilities discussed throughout this section.

\paragraph{Build and debug.} In multi-turn agentic tasks, one performant strategy we observe is GPT being deployed as a builder, making use of its impressive performance in agentic coding, with Opus being deployed at crucial moments to reverify quality and expose vulnerabilities, making use of Opus' expertise in debugging. When \textsc{Fugu-Ultra} was tasked to build a PyPI server hosting a vectorops package in Terminal Bench 2.1, \textsc{Fugu-Ultra} first deployed GPT-5.5 to build the server and confirm the server's reachability after it finished. After GPT finished the build, Opus was then deployed to enumerate all concrete risks in the GPT's implementation, for which Opus noticed 1) GPT used a plain static http.server rather than a pypiserver, 2) the hand-built wheel was built via zipfile, which was fragile and 3) GPT had failed to properly manage the Debian-slim environment, which would lead to several commands failing. Noticing these issues, Opus then discovered GPT's reachability check was erroneous and had come from an orphaned http.server. Relaying these findings back to GPT enabled GPT to complete the build successfully.  

For \textsc{Fugu}, we similarly observe select application of Opus's debugging expertise to expose and correct vulnerabilities in GPT's build. In Terminal Bench 2.1's merge-diff-arc-agi-task, \textsc{Fugu} allowed GPT to build out the scaffold, fetch the two bundles into the branches, and perform the merge in algo.py. At the point GPT recieved the merge conflict back from the environment, \textsc{Fugu} then swapped in Opus, which then both rechecked the scaffold and resolved the merge initiated by GPT, which then revealed the required input-output mapping as originating from neither branch, prompting Opus to then correctly infer the solution from first principles. This per-step alternation between GPT-as-builder and then Opus-as-debugger was critical to allowing \textsc{Fugu} to access performance beyond either model through merging their fine-grained and complementary capabilities.

In other agentic coding tasks, we observe \textsc{Fugu-Ultra} alternating Opus into the lead engineer role with GPT probing for weaknesses and offering alternate perspectives. One example of this was during \textsc{Fugu-Ultra}'s solution to a gravitational/teleport instance in SWE Bench Pro, where the model needs to resolve a failed validation error code when a user adds an additional OTP device. \textsc{Fugu-Ultra} deployed Opus to understand the issue and propose a fix, to which Opus tracked the server registration and validation all the way into the TOTP library itself, ultimately leading to a dead end. At this point, \textsc{Fugu-Ultra} then called GPT to re-examine the issue from a clean slate, to which GPT noticed the issue was not a server-side issue at all but a client-side concurrency bug, for which the server error was merely a symptom of a larger, structural bug. GPT then relayed this information back to Opus which duly changed course and identified the proper fix, which was to introduce a single, shared ContextReader to prevent the concurrency bug.

\paragraph{Bringing in a specialist.} Another strategy we observed delivering strong performance was the select usage of additional models when highly specific knowledge was required. To take another example from Terminal Bench 2.1, an interesting instance of this was during \textsc{Fugu-Ultra}'s solution to the feal-differential-cryptanalysis problem. In this problem, the model is tasked to write an attack that recovers a key from a FEAL-like encryption function. Making use of Opus-4.8's expertise in cybersecurity and software engineering, \textsc{Fugu-Ultra} started the task with Opus building a first-pass chosen-plaintext attack. After Opus was satisfied with its attack, \textsc{Fugu-Ultra} then made use of GPT-5.5's mathematical expertise, tasking it to act specifically as a math specialist to re-derive the entire attack from first principles, and in particular trace through the cipher bit-by-bit to derive the differential constant necessary for the attack to succeed and recover the required round subkey. This example illustrates \textsc{Fugu-Ultra}'s ability to recognize the need to combine cross-domain expertise, drawing on cybersecurity, engineering, and math, and efficiently leverage that expertise from within its agent team to solve complex, multi-disciplinary problems.

%% file: sections/06_conclusion.tex
We present Sakana Fugu, a family of learned LLM orchestrators that expose multi-agent intelligence through a single model interface.
Motivated by the complementary specializations of frontier models and the growing importance of agentic scaffolds, Fugu models learn to construct query-adaptive workflows over a pool of expert LLM workers.
We introduced two variants: \textsc{Fugu}, which balances performance and latency for everyday interactive use, and \textsc{Fugu-Ultra}, which prioritizes answer quality for difficult multi-step tasks.
Building on the Trinity and Conductor frameworks, and extending them with production-oriented design choices, Sakana Fugu demonstrates that model orchestration can serve as a practical scaling axis for frontier AI capabilities.
Our evaluations and early user experiences suggest that learned coordination can achieve strong performance on challenging coding, reasoning, and agentic tasks while providing a simple interface to a complex multi-agent system.

Looking ahead, we believe learned orchestration will become increasingly important as the ecosystem of frontier models continues to diversify.
Because Sakana Fugu composes models at the behavioral level rather than the parameter level, it can incorporate new worker models as they become available, adapt to different user constraints, and support a wider range of deployment and privacy requirements.
More broadly, Sakana Fugu points toward a view of AI progress in which capabilities arise not only from scaling individual models but also from learning how specialized models can collaborate.
We hope this report encourages further research into collective intelligence, query-adaptive scaffolds, and production-ready multi-agent systems.

%% file: sections/appendix.tex
\appendix

\section{Additional Evaluation Configuration Details}\label{app: eval details}

\paragraph{SWE Bench Pro.} We set the max turns to 1000, effectively disabling any turn cap and allowing the model to work until completion. We report results using Mini-Swe-Agent. We additionally tested our model through EvalScope's default configuration (v1.8.1) with resource limit set to 32GB RAM, 4 CPU, and using the toolcalling harness, and found consistent results. All baseline results are provider-reported.

\paragraph{Terminal Bench 2.1.} Results are obtained using the latest EvalScope(v 1.8.1) default configuration with 500 max turns.
We use the Terminus 2 harness and use either provider-reported results (when results are available for Terminus 2) or the Terminal Bench leaderboard from tbench.ai. 

\paragraph{LiveCodeBench v6.} We run the evaluation on the latest release v6 version, May 2023 to the April 2025, which consists of 1055 questions. We patched EvalScope's execution utility to add a buffer attribute to sys.stdin, allowing models to read input via sys.stdin.buffer.read(). This prevents legitimate solutions raising attribute errors. We report all baseline model scores from vals.ai \citep{valsai_livecodebench}. 

\paragraph{LiveCodeBench Pro.} We run the evaluation text-only and without tools. Occasional LiveCodeBench Pro samples include image URLs in the footnotes, however we choose not to pass any web fetch tools to the models to allow them to fetch these URLs, relying only on the problem description in the main body of the problem. We use the 2025 quarter 2 split, which is the most recent publicly available and gradable split. We run all baselines with five retries on timeouts or max token exhaustion. 

\paragraph{Humanity's Last Exam.} We run the evaluation with the full 2500 samples, including multimodal samples, and without providing tools to the models. All baseline results are either provider-reported or collected from Artificial Analysis.

\paragraph{CharXiv Reasoning.} We use gpt-4o as the LLM judge. All baseline results are provider-reported, with the exception of Opus-4.8, which was self-computed.

\paragraph{GPQA-Diamond.} We use the default configuration in EvalScope (v1.8.1). All baseline results are provider-reported. 

\paragraph{SciCode.} We follow the Artificial Analysis implementation, using Inspect AI's default implementation with background provided, and report the final score as the resolve rate over the canonical 288 subproblems. We found that certain SciCode test cases can fail legitimate solutions due to outdated package versioning, requiring us to version bump numpy, scipy, and sympy. All baseline results are taken from Artificial Analysis's official leaderboard. 

\paragraph{$\tau^3$ Banking.} We follow the default configuration from $\tau^3$-bench banking setting, with all tool available for information retrieval, reporting pass@4 accuracy with GPT-5.2 (low) as the user simulator. 
All baseline results are taken from the official $\tau$-bench leaderboard.

\paragraph{Long Context Reasoning.} We follow the default setup as in Artificial Analysis \citep{artificialanalysis2025lcr} with zero few shot examples, using Qwen3 235B A22B 2507 Non-Reasoning as an equality checker and a two hour timeout setting. All baseline results are taken from Artificial Analysis's leaderboard. 

\paragraph{MRCR v2.} We run the benchmark using the standard 8-needle retrieval up to 128k context length. All baseline results are provider-reported.

References to Mythos / Fable 5 and Mythos preview results are collected from official Anthropic-reported material wherever possible, and from official leaderboards and third-party services (such as tbench.ai, Artificial Analysis, Benchmarklist, and vals.ai) when Anthropic-provided scores are unavailable.

\section{More Experiments}\label{app:more_exp}

\subsection{One-shot Rubik's Cube Solver}
We test whether a model can synthesize a correct, near-optimal combinatorial solver from scratch in a single attempt. Each model is prompted once to write a function solve(facelet) that returns a WCA move string, restricted to the Python standard library (no off-the-shelf cube-solving package), so the model must implement both the cube model and the search itself. The emitted code is executed on a frozen, held-out set of 300 scrambled cubes and checked by an independent cube engine; because solving runs as local CPU, no further model tokens are spent at evaluation time. Beyond the shared scaffold, this task carries an absolute quality reference: God's number is 20, giving an absolute upper bound on optimal solution length, and we report solve rate, mean solution length in the half-turn metric (HTM), and mean solver runtime per cube.

\begin{table}[!ht]
\centering
\caption{Rubik's cube solver synthesis on 300 held-out scrambles. Both \textsc{Fugu} models and one frontier baseline (Model~A) solve every cube; the other two frontier baselines crash before solving a single one. \textsc{Fugu-Ultra} produces the shortest solutions of any model (mean 19.72 HTM, optimal $=20$) and never returns a longer solution than Model~A on any cube; \textsc{Fugu} trades about one extra move for a far faster solver (1.9\,s per cube, roughly 35$\times$ faster than the deep solvers' $\sim$70\,s). HTM, half-turn metric; ``mean time'' is the solver's per-cube runtime.}
\begin{tabular}{lccc}
\hline
\textbf{Model} & \textbf{Solve rate} & \textbf{Mean HTM} & \textbf{Mean time} \\
\hline
\textsc{\textbf{Fugu-Ultra}} & 300/300 & \textbf{19.72} & 72.6\,s \\
Model A    & 300/300 & 19.76 & 67.2\,s \\
\textsc{\textbf{Fugu}}       & 300/300 & 21.15 & \textbf{1.9\,s} \\
Model B    & 0/300   & ---   & --- \\
Model C    & 0/300   & ---   & --- \\
\hline
\end{tabular}
\label{tab:rubik}
\end{table}

The decisive result is the reliability of the synthesized code (Table \ref{tab:rubik}). Both Fugu models, \textsc{Fugu} and \textsc{Fugu-Ultra}, produced solvers that ran and solved every cube. Among the three frontier baselines, one (Model A) also solved every cube, but the other two shipped solvers that crashed before solving a single cube. Producing a solver that actually runs is exactly where two of the three frontier baselines fail and where both \textsc{Fugu} models do not.

On solution length, \textsc{Fugu-Ultra} is the strongest model: it produces the shortest solutions of any model (mean 19.72 HTM, within one move of the optimal bound of 20, and ahead of Model~A's 19.76), and on no cube does it return a longer solution than Model~A (7 wins, 293 ties, 0 losses across all 300 cubes). \textsc{Fugu} trades about one extra move (21.15 HTM) for a far faster solver, solving each cube in \textbf{1.9,s}, roughly 35x faster than the deep solvers' ~70,s. Both \textsc{Fugu} models achieve this while solving every cube, exactly where two of the three frontier baselines fail to run at all.

\subsection{Blindfold Chess Gameplay}
Blindfold chess is a stress test for one specific ability: to hold a complex, changing state in mind, and to think about it for a long time. 
Each model is given a fixed opening as a list of moves (never as a board in ASCII or FEN string) and continues from there entirely from memory. 
On each turn, it receives only the opponent's last move in coordinate notation and must return its own, within a single continuous session.
We do not provide a list of legal moves available, and models are expected to find one on its own.
In addition, no external record is maintained in the conversation history.
We expect the model to track the entire position on its own by internally maintaining a record of every capture, castle, and promotion over a game that may span dozens of moves.
It is important to note that in a game of chess, a single forgotten or misplaced piece is typically sufficient to lose.

\begin{figure}[!t]
    \centering
    \includegraphics[width=\linewidth]{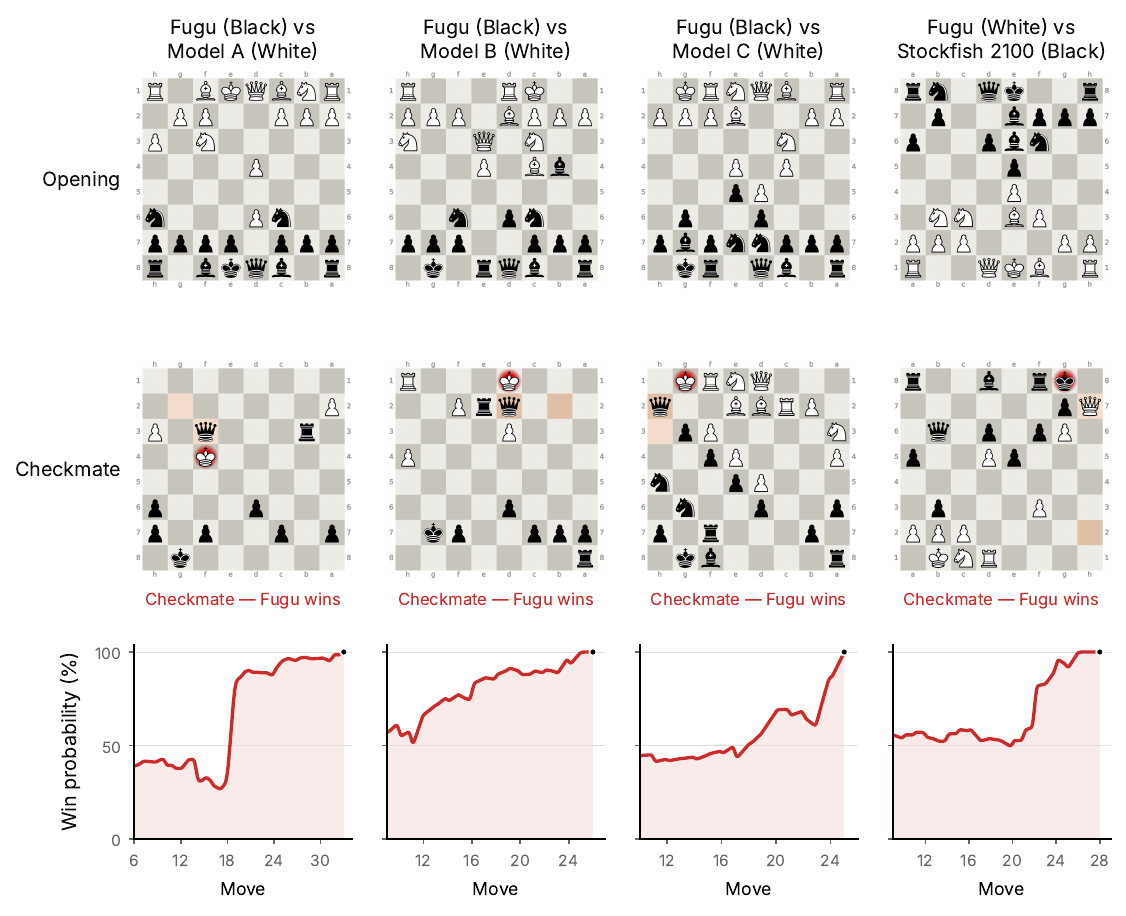}
    \caption{
    Four representative blindfold chess games: \textsc{Fugu} against each of the three anonymized frontier baselines (Model~A, Model~B, Model~C) and against an expert-strength Stockfish~18 (${\sim}2100$~Elo).  
    (Top) the position where agent play begins (\emph{Opening});
    (Center) the final checkmate position (\emph{Checkmate}); and (Bottom)
    \textsc{Fugu}'s win probability over the course of the game (\emph{Win probability}, from Stockfish and the black dot marks the checkmate).
    Neither side ever sees a board in ASCII or FEN, games are played from memory in coordinate notation, and \textsc{Fugu}'s pieces are at the bottom of each chess board. 
    \textsc{Fugu} wins all four and plays them more accurately than its opponent (lower centipawn loss, with no blunders or mistakes of its own). 
    These are selected, illustrative games, not an aggregate or a win rate.}
    \label{fig:blindfold}
\end{figure}

We run the same protocol against the three frontier baselines Model~A, Model~B, Model~C and against Stockfish~18~\citep{stockfish18}, whose playing strength we cap to an expert ${\sim}2100$~Elo
\footnote{Win probabilities and per-move quality labels are obtained from a Stockfish~18 analysis at search depth~8 with 10 principal variations; the engine's centipawn score is mapped to a win percentage through the Lichess win percentage logistic~\citep{lichess_accuracy}, and the move labels (\emph{best}, \emph{excellent}, \emph{good}, \emph{inaccuracy}, \emph{mistake}, \emph{blunder}) follow standard centipawn-loss thresholds. 
As an opponent, Stockfish~18 has its playing strength limited to the stated Elo through the built-in UCI strength limiter of the engine.}.
Figure~\ref{fig:blindfold} shows four representative games: \textsc{Fugu} against each frontier baseline and against an expert-strength Stockfish (2100-Elo), each with its opening, its final checkmate, and the win-probability curve between them.
We provide a complete move history of the games in Appendix~\ref{app:blindfold-games}. 
The central observation is that \textsc{Fugu} sustains accurate play over the course of a full game.
In all four games, it commits no blunders and no mistakes (only occasional minor inaccuracies) under the engine's labeling, whereas each opponent commits at least one mistake or blunder, which proves decisive. 
Its mean average centipawn loss (ACPL) is accordingly lower than its opponent's in every game (${\approx}\,18$--$30$ for \textsc{Fugu} versus ${\approx}\,46$--$72$ for the opponents). 

The most demanding case is the 2100-Elo Stockfish against which \textsc{Fugu} (as White) plays a game that the engine's own analysis finds entirely free of inaccuracies. 
The win-probability trajectories further indicate that these are not one-sided games, against Model~A, \textsc{Fugu} enters the agent-played phase at a slight disadvantage and recovers to a winning position rather than converting a pre-existing opening advantage. 
We present these as illustrative examples rather than as an estimate of the win rate. 
The main objective of this experiment is to isolate the capability of the models to do accurate calculation over a long horizon without an external record of state and to show that \textsc{Fugu} maintains it consistently across opponents.

\textbf{Reproducibility.} 
Every game uses the same two prompt templates, reproduced in Listing~\ref{lst:blindfold}. 
A one-time \emph{setup} prompt names the side to move and gives the opening as a list of coordinate moves, from then on, each turn sends only the opponent's latest move, in the same session, so the board is never restated.
The task does not use an agentic scaffold; every model is queried directly through its bare API under an identical prompt and harness, so the measurement reflects the underlying model rather than the surrounding tooling.

\begin{lstlisting}[float=t, caption={
Blindfold chess prompts. 
The setup prompt is sent once and thereafter each turn supplies only the opponent's move in coordinate notation within the same session, so the model must track the board itself.
}, label={lst:blindfold}, basicstyle=\ttfamily\scriptsize, breaklines=true, frame=single, keywordstyle=\color{black}]
# Setup prompt (sent once, on the first turn)
You are playing a chess game and you are playing with black pieces. Current move history is 1. e2e4 e7e5 2. d2d4 e5d4 3. d1d4 b8c6 4. d4e3 g8f6 5. b1c3 f8b4 6. c1d2 e8g8 7. e1c1 f8e8 8. f1c4 d7d6 9. g1h3. What is your move?

# Per-move prompt (every later turn, same session): opponent's move in UCI only
c1e3
\end{lstlisting}

\subsection{Online Sequential Stock Trading}
We ask whether agents can trade profitably online, where future information is unavailable and each decision permanently shapes the portfolio
\footnote{This benchmark uses a single anonymized equity over one historical 50-week window and is intended to compare sequential, no-look-ahead decision-making rather than to establish generalizable trading performance. Results may not transfer to other assets, time periods, or live markets.}.
The method extends the next-action prediction paradigm, studied in sequential recommendation~\citep{sasrec} and agentic recommendation~\citep{agentic_recsys}, into an agentic trading task requiring the agent to both anticipate future dynamics and commit to irreversible decisions.
Built on the Apple Inc.\ Stock Price Dataset from Kaggle ~\citep{apple_stock_1980_2021}, the experiment follows the causal, no-look-ahead principle of LSTM-based financial prediction ~\citep{lstm_stock_pred}. But where such work uses batch walk-forward evaluation, we adopt a fully online, per-step setting: the agent observes only past prices for a single asset (STOCK\_X), commits to an action, and sees the true next value only afterward, before acting again with accumulated feedback. We disable web search in addition to  anonymizing and normalizing the data into weekly observations, so performance reflects genuine reasoning rather than memorized ticker knowledge.

Each agent begins with \$10,000 and proceeds through an identical 50-week pipeline. 
At every step it observes only current and past weekly market data (opening, high, low, and closing prices, volume, 1/4/12-week returns, 4/12/26-week moving averages, 12-week volatility, and 26-week drawdown), together with its current portfolio state and its own prior feedback.
Each agent maintains a persistent memory log it updates after each step, and chooses both a direction (buy, hold, or sell) and a sizing (25\%, 50\%, or 100\% of available cash or held shares). After each action the next week's price is revealed and the portfolio is marked to market, so the agent must continually revise its strategy from realized outcomes rather than peeking ahead. 
All frontier models are run under this identical setup, with the same prompts, action space, and evaluation harness, to ensure a fair comparison.

\begin{figure}[!hb]
    \centering
    \includegraphics[width=0.8\linewidth]{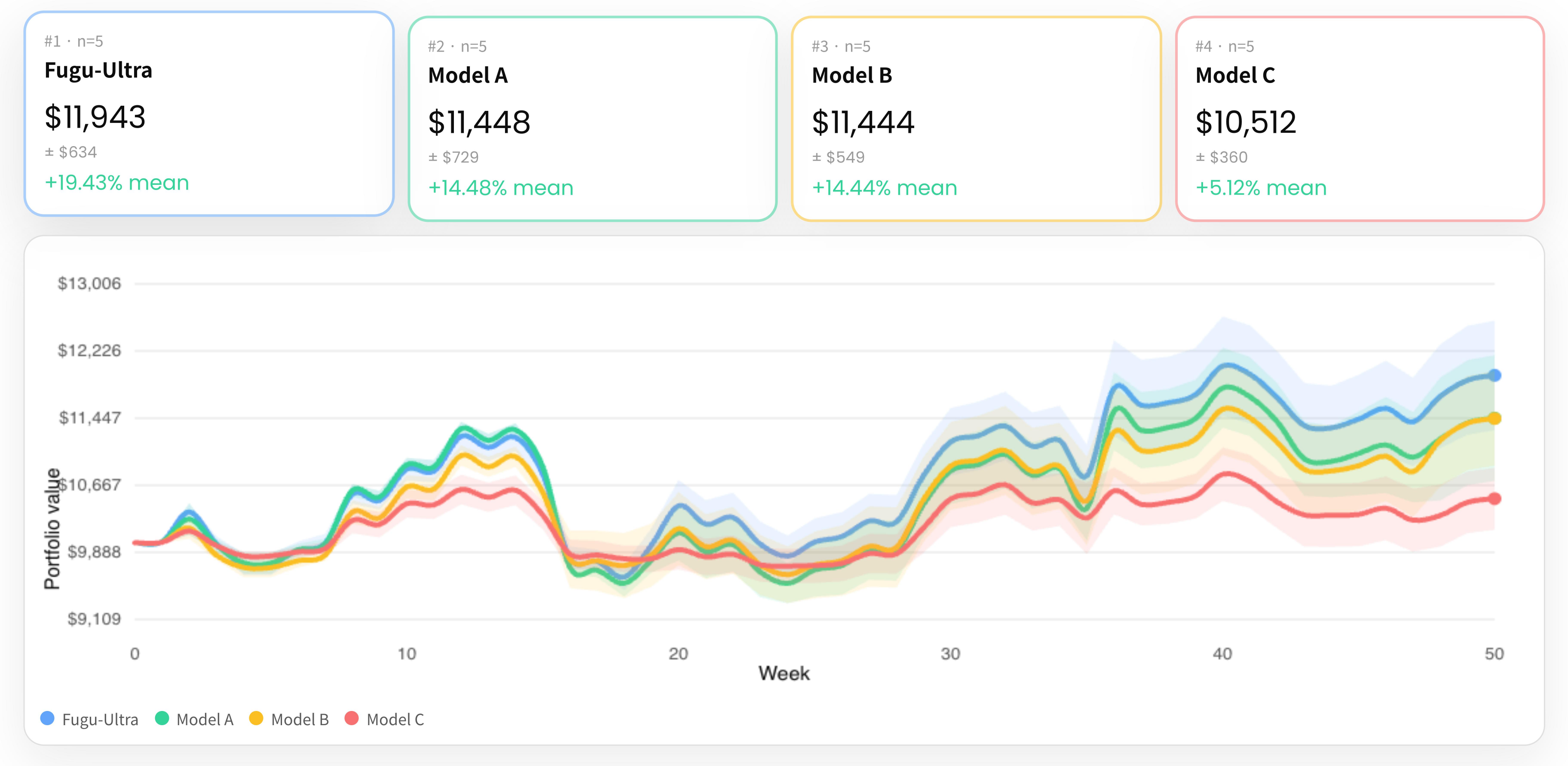}
    \caption{Final portfolio value over the 50-week online trading run on STOCK\_X, starting from \$10{,}000. \textsc{Fugu-Ultra} (blue) reaches
    \$11{,}943.22 ± \$633.86 (+19.43\% mean return), while the other frontier models reached their return less than +15\%. 
    }
    \label{fig:trading}
\end{figure}

Across five runs of the identical 50-week pipeline, \textsc{Fugu-Ultra} produced the strongest result, growing the portfolio to \textbf{\$11{,}943.22 ± \$633.86 (+19.43\% mean return)}, while the other frontier models reached their return less than +15\%. As shown in Figure \ref{fig:trading}, \textsc{Fugu-Ultra} compounds gains more effectively over the sequence, whereas the other frontier models stay their trades less strategic. The gap suggests that \textsc{Fugu-Ultra} is better at translating noisy market signals into well-timed, appropriately-sized trades and at adapting its position from week-to-week feedback, rather than defaulting to passive or overly conservative behavior.

\section{Blindfold Chess: Full Game Records}\label{app:blindfold-games}
For reproducibility, we give the complete move record (standard algebraic notation) and the final board position (FEN) for each of the four games in Figure~\ref{fig:blindfold}. 
Each record is split into the fixed \emph{opening} (the shared book sequence, supplied to both models as a list of moves) and the subsequent \emph{blindfold play} (generated move-by-move from memory, with no board or legal-move list shown).

\paragraph{\textsc{Fugu} (Black) vs Model A (White).} Result: 0--1 (\textsc{Fugu} wins).
\begin{lstlisting}[breaklines=true, basicstyle=\ttfamily\scriptsize, frame=single, keywordstyle=\color{black}]
--- Opening (fixed book moves, given to both models) ---
1. e4 Nc6 2. Nf3 Nf6 3. e5 Ng4 4. d4 d6 5. h3 Nh6 6. exd6

--- Blindfold play (from memory; no board or legal moves shown) ---
6... exd6 7. Bxh6 gxh6 8. Nc3 Bg7 9. Bb5 O-O 10. O-O Bd7 11. Re1 Re8 12. Rxe8+ Qxe8 13. Bxc6 Bxc6 14. d5 Bxc3 15. dxc6 Bxb2 16. Rb1 Ba3 17. cxb7 Rb8 18. Qd3 Bc5 19. Qa3 Bxa3 20. Rb3 Bc5 21. c3 Qe6 22. Nd4 Bxd4 23. cxd4 Qe1+ 24. Kh2 Qxf2 25. Rb2 Qxb2 26. Kg3 Qxd4 27. Kf3 Rxb7 28. Kg3 Qe5+ 29. Kf3 Qf5+ 30. Kg3 Qg5+ 31. Kf3 Rb3+ 32. Ke4 Qxg2+ 33. Kf4 Qf3# 0-1

Final FEN: 6k1/p1p2p1p/3p3p/8/5K2/1r3q1P/P7/8 w - - 2 34
\end{lstlisting}

\paragraph{\textsc{Fugu} (Black) vs Model B (White).} Result: 0--1 (\textsc{Fugu} wins).
\begin{lstlisting}[breaklines=true, basicstyle=\ttfamily\scriptsize, frame=single, keywordstyle=\color{black}]
--- Opening (fixed book moves, given to both models) ---
1. e4 e5 2. d4 exd4 3. Qxd4 Nc6 4. Qe3 Nf6 5. Nc3 Bb4 6. Bd2 O-O 7. O-O-O Re8 8. Bc4 d6 9. Nh3

--- Blindfold play (from memory; no board or legal moves shown) ---
9... Bxc3 10. Bxc3 Nxe4 11. Qf4 Nxc3 12. bxc3 Be6 13. Bd3 Ne5 14. g4 Bxg4 15. Rdg1 Nxd3+ 16. cxd3 Bxh3 17. Qg3 Qf6 18. Qxh3 Qxc3+ 19. Kd1 Qa1+ 20. Kd2 Qxa2+ 21. Kd1 Qb1+ 22. Kd2 Qb2+ 23. Kd1 Re2 24. Qxh7+ Kxh7 25. Rxg7+ Kxg7 26. h4 Qd2# 0-1

Final FEN: r7/ppp2pk1/3p4/8/7P/3P4/3qrP2/3K3R w - - 1 27
\end{lstlisting}

\paragraph{\textsc{Fugu} (Black) vs Model C (White).} Result: 0--1 (\textsc{Fugu} wins).
\begin{lstlisting}[breaklines=true, basicstyle=\ttfamily\scriptsize, frame=single, keywordstyle=\color{black}]
--- Opening (fixed book moves, given to both models) ---
1. d4 Nf6 2. c4 g6 3. Nc3 Bg7 4. e4 d6 5. Nf3 O-O 6. Be2 e5 7. O-O Nc6 8. d5 Ne7 9. Ne1 Nd7

--- Blindfold play (from memory; no board or legal moves shown) ---
10. Be3 f5 11. f3 f4 12. Bf2 g5 13. Nd3 Nf6 14. c5 Ng6 15. cxd6 cxd6 16. Rc1 Rf7 17. Nb5 Bf8 18. a4 g4 19. Rc2 g3 20. Be1 Nh5 21. h3 a6 22. Na3 Bxh3 23. gxh3 Qd7 24. Bd2 Qxh3 25. Ne1 Qh2# 0-1

Final FEN: r4bk1/1p3r1p/p2p2n1/3Pp2n/P3Pp2/N4Pp1/1PRBB2q/3QNRK1 w - - 2 26
\end{lstlisting}

\paragraph{\textsc{Fugu} (White) vs Stockfish 2100 (Black).} Result: 1--0 (\textsc{Fugu} wins).
\begin{lstlisting}[breaklines=true, basicstyle=\ttfamily\scriptsize, frame=single, keywordstyle=\color{black}]
--- Opening (fixed book moves, given to both models) ---
1. e4 c5 2. Nf3 d6 3. d4 cxd4 4. Nxd4 Nf6 5. Nc3 a6 6. Be3 e5 7. Nb3 Be6 8. f3 Be7

--- Blindfold play (from memory; no board or legal moves shown) ---
9. Qd2 O-O 10. O-O-O b5 11. g4 Qc7 12. h4 Nbd7 13. g5 Nh5 14. Kb1 Ng3 15. Rg1 b4 16. Nd5 Bxd5 17. exd5 Nxf1 18. Rgxf1 a5 19. Nc1 f6 20. g6 hxg6 21. Rg1 g5 22. hxg5 Nb6 23. Bxb6 Qxb6 24. g6 Bd8 25. Rh1 b3 26. Rh8+ Kxh8 27. Qh2+ Kg8 28. Qh7# 1-0

Final FEN: r2b1rk1/6pQ/1q1p1pP1/p2Pp3/8/1p3P2/PPP5/1KNR4 b - - 3 28
\end{lstlisting}